\documentclass[12pt]{article}
\usepackage[margin=1in]{geometry}
\usepackage{amsmath,amssymb,amsfonts}
\usepackage{graphicx}
\graphicspath{{figures/}}
\usepackage{textcomp}
\usepackage{url}
\usepackage{hyperref}
\usepackage{subcaption}
\usepackage{bm}
\usepackage{booktabs}
\usepackage{multirow}
\usepackage{adjustbox}
\usepackage{xcolor}
\usepackage[numbers,sort&compress]{natbib}
\usepackage[affil-it]{authblk}

\usepackage{microtype}

\definecolor{medvaeblue}{HTML}{1565C0}
\definecolor{sdvaered}{HTML}{C62828}
\newcommand{\hledit}[1]{#1}
\newcommand{\medvaesr}{MedVAE SR}
\newcommand{\sdvaesr}{SD-VAE SR}
\newcommand{\medvaeae}{MedVAE-AE}
\newcommand{\sdvaeae}{SD-VAE-AE}

\begin{document}

\title{Domain-Specific Latent Representations Improve the Fidelity of Diffusion-Based Medical Image Super-Resolution}

\author[1]{Sebastian Cajas}
\author[8]{Ashaba Judith}
\author[1,11,12]{Rahul Gorijavolu}
\author[6]{Sahil Kapadia}
\author[2,3]{Hillary Clinton Kasimbazi}
\author[7,8,9]{Leo Kinyera}
\author[13]{Emmanuel Paul Kwesiga}
\author[1]{Sri Sri Jaithra Varma Manthena}
\author[1,10]{Luis Filipe Nakayama}
\author[8]{Ninsiima Doreen}
\author[1,4,5]{Leo Anthony Celi}

\affil[1]{MIT Critical Data, Massachusetts Institute of Technology, Cambridge, MA, USA.}
\affil[2]{Department of Radiology and Radiotherapy, Makerere University, Uganda.}
\affil[3]{Department of Radiotherapy, Uganda Cancer Institute, Uganda.}
\affil[4]{Division of Pulmonary, Critical Care and Sleep Medicine, Beth Israel Deaconess Medical Center, Boston, MA, USA.}
\affil[5]{Department of Biostatistics, Harvard T.H.\ Chan School of Public Health, Boston, MA, USA.}
\affil[6]{University of North Carolina at Chapel Hill, Chapel Hill, NC, USA.}
\affil[7]{Department of Radiology and Radiological Science, School of Medicine, Johns Hopkins University, Baltimore, MD, USA.}
\affil[8]{Department of Biomedical Engineering, Mbarara University of Science and Technology, Uganda.}
\affil[9]{Department of Public Health Sciences, Cavendish University, Uganda.}
\affil[10]{Department of Ophthalmology, Federal University of S\~{a}o Paulo, Brazil.}
\affil[11]{School of Medicine, Johns Hopkins University, Baltimore, MD, USA.}
\affil[12]{Department of Biomedical Engineering, Johns Hopkins University, Baltimore, MD, USA.}
\affil[13]{Department of Applied Natural Sciences, Technical University of Applied Sciences L\"{u}beck (TH L\"{u}beck), L\"{u}beck, Germany.}
\date{}

\maketitle

\vspace{1em}
\begin{abstract}
Latent diffusion models for medical image super-resolution universally inherit
variational autoencoders designed for natural photographs.
We show that this default choice\,--\,not the diffusion architecture\,--\,is the
dominant constraint on reconstruction quality.
In a controlled experiment holding all other pipeline components fixed,
replacing the generic Stable Diffusion VAE with MedVAE, a domain-specific
autoencoder pretrained on $>1.6$M medical images, yields $+2.91$ to $+3.29$~dB
PSNR improvement across knee MRI, brain MRI, and chest X-ray
($n=1{,}820$; Cohen's $d = 1.37$--$1.86$, all $p < 10^{-20}$, Wilcoxon signed-rank).
Wavelet decomposition localises the advantage to the finest spatial frequency
bands encoding anatomically relevant fine structure.
Ablations across inference schedules, prediction targets, and generative
architectures confirm the gap is stable within $\pm 0.15$~dB, while hallucination
rates remain comparable between methods (Cohen's $h < 0.02$ across all datasets),
establishing that reconstruction
fidelity and generative hallucination are governed by independent pipeline components.
These results provide a practical screening criterion: autoencoder reconstruction
quality, measurable without diffusion training, predicts downstream SR performance
($R^2 = 0.67$), suggesting that domain-specific VAE selection should precede
diffusion architecture search.
Code and trained model weights are publicly available at \url{https://github.com/sebasmos/latent-sr}.
\end{abstract}


\section{Introduction}

Medical imaging quality varies enormously across the globe, with low- and
middle-income countries (LMICs) disproportionately reliant on low-field MRI
systems that produce images of inferior spatial resolution compared to the
1.5\,T and 3\,T scanners standard in high-income
settings~\citep{muraliBringingMRILow2024, arnoldLowFieldMRI2023}.
An estimated two-thirds of the world's population lacks access to MRI, and in
regions such as sub-Saharan Africa, the majority of installed devices operate
at field strengths below 0.3\,T~\citep{anazodoFrameworkSustainableMRI2022, jonesLowFieldHighImpact2025}.
Super-resolution (SR)\,--\,computational techniques that reconstruct
higher-resolution images from lower-resolution inputs\,--\,has emerged as a
promising approach to bridge this gap, with recent diffusion model--based
methods demonstrating substantial improvements in image fidelity metrics
such as Peak Signal-to-Noise Ratio (PSNR), Structural Similarity Index (SSIM),
and Learned Perceptual Image Patch Similarity
(LPIPS)~\citep{dubeyTemporalSpatialSR2024, wangInverseSR2023}.
However, a critical concern remains: generative SR models, particularly those
based on latent diffusion, may introduce hallucinated anatomical details that
appear visually convincing but have no correspondence to the true underlying
anatomy~\citep{kcSFRC2026, renHallucinationScore2025}.
This tension between perceptual quality and diagnostic fidelity demands
rigorous evaluation frameworks that go beyond downstream task performance, which risks entanglement with shortcut features inherent to the classification model itself.

High-resolution medical images underpin clinical diagnosis: fine anatomical
structure in an MRI scan distinguishes tumour margin from oedema; sharp
vascular detail in a chest radiograph separates consolidation from artefact.
Higher magnetic field strengths deliver proportionally finer anatomical detail
and improved grey--white matter contrast~\citep{jonesNeuroimaging3TVs2021, laddProsConsUltrahighfield2018},
yet the associated cost, safety infrastructure, and limited global availability
make SR an important clinical target~\citep{muraliBringingMRILow2024}.
Deep learning has substantially advanced SR quality~\hledit{\citep{wang2021, yueImageSuperresolutionTechniques2016, anwarDeepJourneySuper2020}}.
Early MRI SR relied on interpolation, including zero-filled k-space methods~\citep{duReductionPartialvolumeArtifacts1994}
and voxel-shift interslice reconstruction~\citep{greenspanMRIIntersliceReconstruction2002};
these provided only modest SNR improvements~\citep{mahmoudzadehInterpolationbasedSuperresolutionReconstruction2014}.
CNN-based approaches, including 3D-SRCNN~\citep{phamMultiscaleBrainMRI2019},
DCSRN~\citep{chenBrainMRISuper2018, masutaniDeepLearningSingleFrame2020},
and GAN-enhanced variants~\citep{chenEfficientAccurateMRI2018, wangESRGANEnhancedSuperResolution2018},
substantially improved reconstruction quality, followed by recent diffusion-based methods
for low-field and clinical MRI~\citep{deleeuwdenbouterDeepLearningbasedSingle2022, baljerUltraLowFieldPaediatricMRI2025};
yet models trained on natural photographs often fail to preserve the statistical regularities
and clinically meaningful fine structure that characterize medical
modalities~\citep{chaudhari2018super, muraliBringingMRILow2024, tianNewClinicalOpportunities2024}.

Latent diffusion models (LDMs)~\citep{rombach2022} represent the current
state of the art for perceptually faithful image synthesis: by performing
iterative denoising in a compressed latent space rather than in raw pixel space,
they generate images of exceptional sharpness and structural coherence.
This capability has motivated their rapid adoption for medical SR~\citep{qiuMedicalImageSuperresolution2023, plengeSuperresolutionMethodsMRI2012}.
LDMs have been applied to brain MRI synthesis~\citep{pinaya2022brain},
accelerated MRI reconstruction by score-based diffusion~\citep{chung2022score,song2023solving},
and cross-modality medical image translation~\citep{ozbey2023unsupervised}.
Concurrent pixel-space SR baselines~\citep{wangESRGANEnhancedSuperResolution2018,liang2021swinir}
and denoising frameworks~\citep{ho2020ddpm,song2021ddim,delbracio2023inversion}
form the comparison set evaluated here.
Yet a fundamental assumption has gone largely unexamined: virtually every
LDM-based medical SR pipeline inherits the variational autoencoder (VAE) from
Stable Diffusion (SD-VAE)~\citep{rombach2022}, a model pretrained on billions of
natural photographs and optimised for photographic fidelity, with no
exposure to the domain statistics of medical imaging~\hledit{\citep{raghuTransfusionUnderstandingTransfer2019}}.
The latent space it defines was never designed to represent the statistical
structure of MRI, computed tomography, or radiography.

Several recent works have proposed domain-specific autoencoders for medical image representation.
MedVAE~\citep{varmaMedVAEEfficientAutomated2025} is a suite of VAEs trained on large-scale medical imaging data across modalities, demonstrating substantially improved reconstruction fidelity over generic natural-image VAEs on MRI, X-ray, and CT.
Similarly, Pinaya et al.~\citep{pinaya2022brain} trained a domain-specific VQ-VAE on brain MRI and showed that domain-adapted latent spaces enable higher-quality generative synthesis.
MedViT~\citep{manzariMedViT2023} and related architectures further demonstrate that medical-domain pretraining improves downstream representation quality.
In the X-ray domain, BioViL-T~\citep{bannur2023biovilT} and RadDINO~\citep{perezgarcia2024raddino} establish that foundation models pretrained on large radiology corpora learn representations that better capture clinically salient structure than ImageNet-pretrained counterparts.
Concurrent pixel-space diffusion SR methods\,--\,SR3~\citep{saharia2022sr3}, which applies iterative refinement in image space, and DiffIR~\citep{xia2023diffir}, which combines a degradation-sensitive prior with a diffusion backbone\,--\,optimise the full pipeline jointly rather than isolating a single component.
Our work is complementary: by holding the diffusion architecture fixed and swapping only the VAE, we isolate the latent space as the independent variable and establish a training-free, modular improvement applicable to any LDM-based SR pipeline.

Despite this body of evidence, the downstream consequence for SR fidelity\,--\,specifically, whether a domain-specific VAE raises the SR quality ceiling\,--\,has not been quantified in a controlled experiment.

This misalignment exposes a fundamental bottleneck.
Medical images are characterised by modality-specific noise textures, narrow
intensity distributions, and anatomical fine structure that differs
substantially from natural-image statistics.
If the VAE encodes these features imperfectly, the diffusion model operates
in a distorted latent manifold, and no amount of additional diffusion training
can recover information the encoder has already discarded.
The VAE reconstruction ceiling, the fidelity achievable by encoding and
decoding without any generative component, is an absolute upper bound
on SR quality (Fig.~\ref{fig:ae_ceiling_corr}).
We hypothesise that a domain-specific VAE pretrained on large-scale medical
imaging data will raise this ceiling and, consequently, yield proportionate
improvements in end-to-end SR performance.

We test this hypothesis with a controlled, single-variable experiment.
Holding the UNet architecture, training objective, noise schedule, and evaluation
protocol fixed, we replace the generic SD-VAE with MedVAE
(\texttt{medvae\_4\_3\_2d})~\citep{varmaMedVAEEfficientAutomated2025}, a domain-specific VAE
pretrained on $>1.6$M medical images spanning radiology, pathology, and
dermatology.
We evaluate on three heterogeneous medical imaging datasets:
knee MRI (MRNet), brain MRI with tumour annotation (BraTS 2023),
and chest X-ray (MIMIC-CXR), covering a $4\times$ SR task and an SR
refinement task, with $n = 1{,}820$ validation images in total.
Rather than assessing SR quality through classification or other downstream
clinical tasks, which risk conflating reconstruction quality with task-specific
model biases, we restrict our evaluation to direct image comparison through
metrics including PSNR and LPIPS that measure pixel-level and perceptual fidelity
against the known ground truth.
Our results demonstrate that domain-specific embeddings yield statistically
significant improvements in reconstruction quality over generic embeddings,
suggesting that encoding clinically relevant visual features in the latent space
is a key factor in producing faithful medical image super-resolution.

Our principal contributions are:
\begin{itemize}
  \item \textbf{A modality-agnostic, training-free ceiling principle for VAE selection.}
        AE reconstruction fidelity (encode-then-decode PSNR, computed without training
        a diffusion model) predicts downstream SR PSNR with Pearson $r = 0.82$ ($R^2 = 0.67$)
        across all method--dataset combinations (Fig.~\ref{fig:ae_ceiling_corr}).
        This training-free, modality-agnostic criterion provides a practical screening tool
        for latent diffusion SR: practitioners can evaluate VAE quality on their target domain
        before committing to expensive diffusion training.
        The principle applies regardless of imaging modality (MRI, X-ray, CT, or other),
        wherever a frozen encoder is inherited from a different domain, and requires
        only a forward encode--decode pass on target-domain images.
        Empirically, replacing SD-VAE with MedVAE raises SR PSNR by $+2.91$ to $+3.29$~dB
        across all three modalities (Cohen's $d = 1.37$--$1.86$; all $p < 10^{-20}$,
        Wilcoxon signed-rank).
  \item \textbf{Mechanistic identification of the VAE as the binding constraint.}
        The MedVAE autoencoder ceiling is $+3.93$ to $\hledit{+6.48}$~dB above
        SD-VAE's ceiling across all datasets, providing strong evidence that latent-space
        fidelity, not diffusion architecture or training objective,
        sets the upper bound on SR quality.
  \item \textbf{A domain-specificity diagnostic via wavelet subband profiling.}
        Three-level Haar wavelet decomposition reveals that domain-specific latent spaces
        concentrate their SR advantage at fine spatial frequencies (HH$_1$: $+1.18$ to $+1.41$~dB)
        while coarse approximation bands are nearly identical ($\leq +0.05$~dB),
        a pattern consistent across all three imaging modalities.
        This predicts that domain-specific VAEs will provide their principal benefit
        at frequencies corresponding to domain-relevant texture in \emph{any} imaging domain,
        and that the subband PSNR profile serves as a training-free diagnostic of VAE domain-match.
  \item \textbf{Independent governance of fidelity and hallucination.}
        Across inference schedules ($T = 50$--$1{,}000$), prediction targets, and
        generative architectures (DDPM vs.\ flow matching), the PSNR gap varies by
        less than $0.15$~dB, confirming that no training-side modification can
        compensate for a mismatched encoder.
        Simultaneously, hallucination rates are comparable (Wilcoxon; Cohen's $h < 0.02$ across all datasets)
        between \medvaesr{} and \sdvaesr{} despite the $+3$~dB PSNR advantage,
        establishing that reconstruction fidelity and generative hallucination
        are governed by distinct pipeline components\,--\,the VAE and the diffusion
        model, respectively\,--\,with direct implications for safety-critical deployment.
\end{itemize}

Together, these results provide strong evidence that the domain-specific latent space is the
primary driver of SR quality in LDMs.

\begin{figure}[htbp]
  \centering
  \makebox[\linewidth][c]{\includegraphics[width=1.25\textwidth]{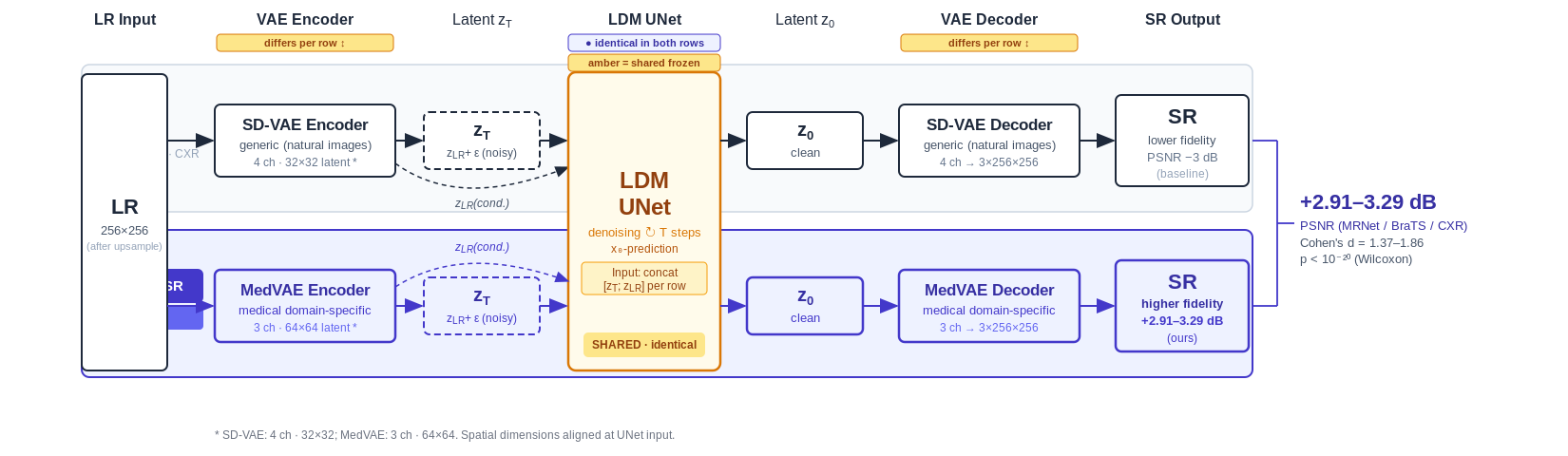}}
  \caption{\textbf{Latent diffusion SR pipeline.}
  Both methods share an identical LDM UNet (purple); only the VAE encoder/decoder differs.
  \textbf{Top:} SD-VAE SR uses a generic VAE trained on natural images (4-channel, $32\times32$ latent).
  \textbf{Bottom:} MedVAE SR (proposed) uses a domain-specific VAE trained on medical images (3-channel, $64\times64$ latent).
  The controlled substitution isolates the VAE's contribution to SR fidelity.
  \label{fig:pipeline}}
\end{figure}

\section{Results}

\subsection{MedVAE SR consistently outperforms SD-VAE SR across three modalities}

\begin{figure}[htbp]
  \centering
  \includegraphics[width=\linewidth]{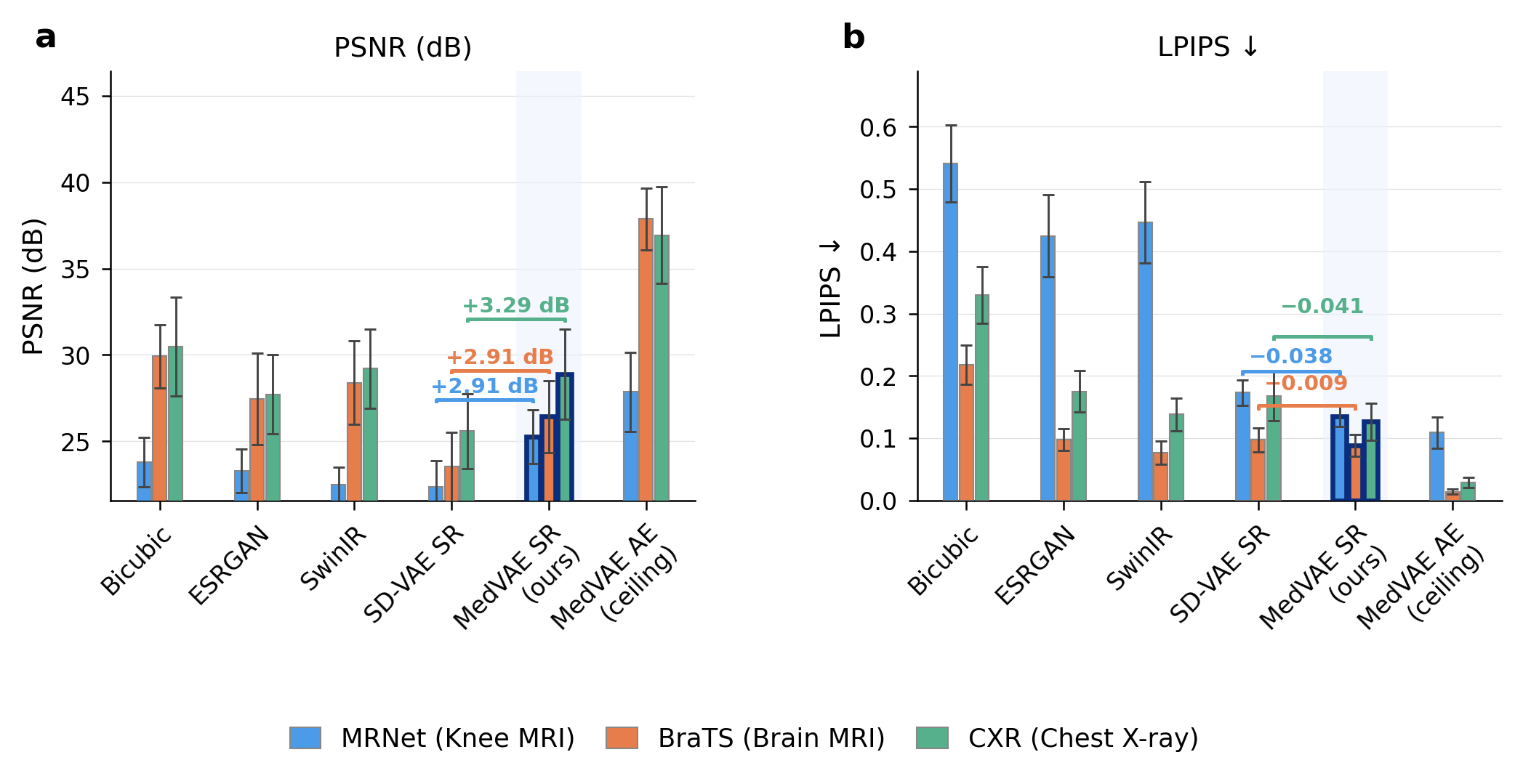}
  \caption{\textbf{SR reconstruction quality across three datasets.}
  \textbf{(a)} PSNR (dB) and \textbf{(b)} LPIPS for all methods on MRNet (knee MRI),
  BraTS (brain MRI), and MIMIC-CXR (chest X-ray).
  Error bars: $\pm1$ standard deviation across validation images.
  Improvement brackets (dark blue) show MedVAE SR vs.\ SD-VAE SR delta:
  $+X$ dB for PSNR with 95\% bootstrap CI, $-\Delta$ for LPIPS (lower = better).
  \medvaesr{} achieves better PSNR than \sdvaesr{} and lower LPIPS than
  bicubic interpolation across all three datasets,
  occupying the Pareto frontier of the quality-perception tradeoff~\citep{blau2018perception}.
  The \medvaeae{} ceiling (light blue, hatched) represents encode-then-decode
  without diffusion and sets the theoretical maximum for each VAE.}
  \label{fig:sr_quality}
\end{figure}

We trained two LDM SR models with identical UNet architectures and training
protocols, differing only in the VAE used to encode and decode images
(Fig.~\ref{fig:pipeline}).
We evaluate on three datasets spanning distinct tasks: MRNet (knee MRI, $2\times$ resolution refinement),
BraTS (brain MRI, $4\times$ SR), and MIMIC-CXR (chest X-ray, $4\times$ SR).
\medvaesr{} consistently outperforms \sdvaesr{} across all three datasets
and all three reference-based metrics (Table~\ref{tab:sr_quality}; Fig.~\ref{fig:sr_quality}).

On the 4$\times$ SR task (BraTS and MIMIC-CXR), \medvaesr{} achieves
$26.42 \pm 2.10$~dB PSNR (BraTS, validation set) and $28.87 \pm 2.62$~dB (CXR),
compared to $23.51 \pm 1.99$~dB and $25.58 \pm 2.17$~dB for \sdvaesr{}
($+2.91$~dB and $+3.29$~dB respectively).
For the MRNet knee MRI dataset, the task is SR refinement rather than
true $4\times$ upsampling: LR images are $2\times$ downsampled then restored to
the original $256\times256$ resolution, testing the model's ability to recover
fine anatomical detail lost during acquisition degradation
(task details in Appendix~\ref{sec:note2}).
This task is complementary to the $4\times$ SR setting on BraTS and CXR:
it isolates high-frequency detail recovery without resolution change,
providing a clinically motivated evaluation of VAE latent quality.
\medvaesr{} achieves $25.26 \pm 1.58$~dB vs.\ $22.34 \pm 1.55$~dB
($+2.91$~dB; 95\% bootstrap CI $[+2.52, +3.30]$, $n=10{,}000$).
All differences are highly significant (Wilcoxon signed-rank,
$p < 10^{-20}$ per dataset; Table~\ref{tab:stats}).

The advantage is not limited to pixel fidelity.
\medvaesr{} also achieves substantially lower perceptual error
(LPIPS$^\ddagger$: $0.088$--$0.135$ vs.\ $0.097$--$0.173$; lower is better),
placing it on the Pareto frontier: better PSNR than SD-VAE SR and
better LPIPS than bicubic interpolation (Table~\ref{tab:sr_quality}).
Bicubic interpolation achieves higher PSNR than diffusion SR methods on BraTS and CXR because it minimises mean squared error (MSE) directly, without introducing any perceptual hallucinations.
MedVAE SR sacrifices some pixel-level fidelity to add perceptual detail, reflected in its substantially lower LPIPS (Appendix Fig.~\ref{fig:supp_pd}).

\begin{table}[h!]
\centering
\caption{\textbf{SR reconstruction quality across three datasets.}
PSNR (dB), MS-SSIM, and LPIPS for all methods.
Bold: best SR method per dataset per metric.
$\dagger$: MRNet is a refinement task (LR\,=\,HR\,=\,256$\times$256);
bicubic copies LR unchanged.
$\ddagger$: BraTS/CXR bicubic PSNR exceeds diffusion SR due to MSE minimisation
vs.\ perceptual quality tradeoff. Bicubic interpolation achieves competitive PSNR
because it avoids hallucination artifacts, but produces blurry outputs with high
LPIPS (low perceptual quality).
$\dagger\dagger$: BraTS results use the validation split ($n=700$); the flow-matching
ablation (Appendix Table~\ref{tab:flow_matching}) uses the test split ($n=720$).}
\label{tab:sr_quality}
\small
\setlength{\tabcolsep}{5pt}
\begin{tabular}{@{}llccc@{}}
\toprule
\textbf{Dataset} & \textbf{Method} & \textbf{PSNR (dB)} & \textbf{MS-SSIM} & \textbf{LPIPS}$^\ddagger$ \\
\midrule
\multirow{5}{*}{MRNet$^\dagger$}
  & Bicubic         & 23.79          & 0.906         & 0.541          \\
  & ESRGAN          & 23.28          & 0.897         & 0.425          \\
  & SwinIR          & 22.48          & 0.882         & 0.446          \\
  & \sdvaesr{}      & 22.34          & 0.870         & 0.173          \\
  & \medvaesr{}$^*$ & \textbf{25.26} & \textbf{0.936}& \textbf{0.135} \\
  \cmidrule(l){2-5}
  & \medvaeae{}$^\S$     & 27.85          & 0.966         & 0.109          \\
\midrule
\multirow{5}{*}{BraTS$^{\ddagger,\dagger\dagger}$}
  & Bicubic         & 29.91          & 0.980         & 0.218          \\
  & ESRGAN          & 27.45          & 0.962         & 0.098          \\
  & SwinIR          & 28.38          & 0.974         & 0.077          \\
  & \sdvaesr{}      & 23.51          & 0.871         & 0.097          \\
  & \medvaesr{}$^*$ & \textbf{26.42} & \textbf{0.936}& \textbf{0.088} \\
  \cmidrule(l){2-5}
  & \medvaeae{}$^\S$     & 37.87          & 1.000         & 0.014          \\
\midrule
\multirow{5}{*}{MIMIC-CXR$^\ddagger$}
  & Bicubic         & 30.47          & 0.977         & 0.330          \\
  & ESRGAN          & 27.71          & 0.951         & 0.175          \\
  & SwinIR          & 29.21          & 0.964         & 0.138          \\
  & \sdvaesr{}      & 25.58          & 0.870         & 0.167          \\
  & \medvaesr{}$^*$ & \textbf{28.87} & \textbf{0.943}& \textbf{0.127} \\
  \cmidrule(l){2-5}
  & \medvaeae{}$^\S$     & 36.93          & 1.000         & 0.029          \\
\bottomrule
\end{tabular}
\par\smallskip{\footnotesize $^*$ Proposed method. $^\S$ MedVAE-AE is an upper-bound oracle (encode then decode with no diffusion); not a competing SR method. $\ddagger$ LPIPS is computed using ImageNet-pretrained features (AlexNet backbone) and may not capture medically salient textures; interpret alongside PSNR and MS-SSIM. PSNR measures reconstruction fidelity while LPIPS captures perceptual quality; these metrics are known to trade off~\citep{blau2018perception}.}
\end{table}

\begin{table}[h!]
\centering
\caption{\textbf{Statistical significance of MedVAE vs.\ SD-VAE SR.}
Per-image Wilcoxon signed-rank test and bootstrap effect sizes ($n_\text{bootstrap}=10{,}000$).}
\label{tab:stats}
\setlength{\tabcolsep}{4pt}
\resizebox{\linewidth}{!}{%
\begin{tabular}{lccccccc}
\toprule
\textbf{Dataset} & \textbf{$n$} & \textbf{MedVAE PSNR} & \textbf{SD-VAE PSNR}
  & \textbf{$\Delta$ PSNR} & \textbf{95\% CI} & \textbf{Cohen's $d$} & \textbf{$p$-value} \\
\midrule
MRNet  & 120   & $25.26 \pm 1.58$ & $22.34 \pm 1.55$ & $+2.91$ & $[+2.52, +3.30]$ & 1.86 & $9.9 \times 10^{-22}$ \\
BraTS  & 700   & $26.42 \pm 2.10$ & $23.51 \pm 1.99$ & $+2.91$ & $[+2.69, +3.12]$ & 1.42 & $7.8 \times 10^{-120}$ \\
CXR    & 1{,}000 & $28.87 \pm 2.62$ & $25.58 \pm 2.17$ & $+3.29$ & $[+3.09, +3.51]$ & 1.37 & $2.3 \times 10^{-117}$ \\
\bottomrule
\end{tabular}}
\end{table}

\begin{figure}[htbp]
  \centering
  \includegraphics[width=0.7\linewidth]{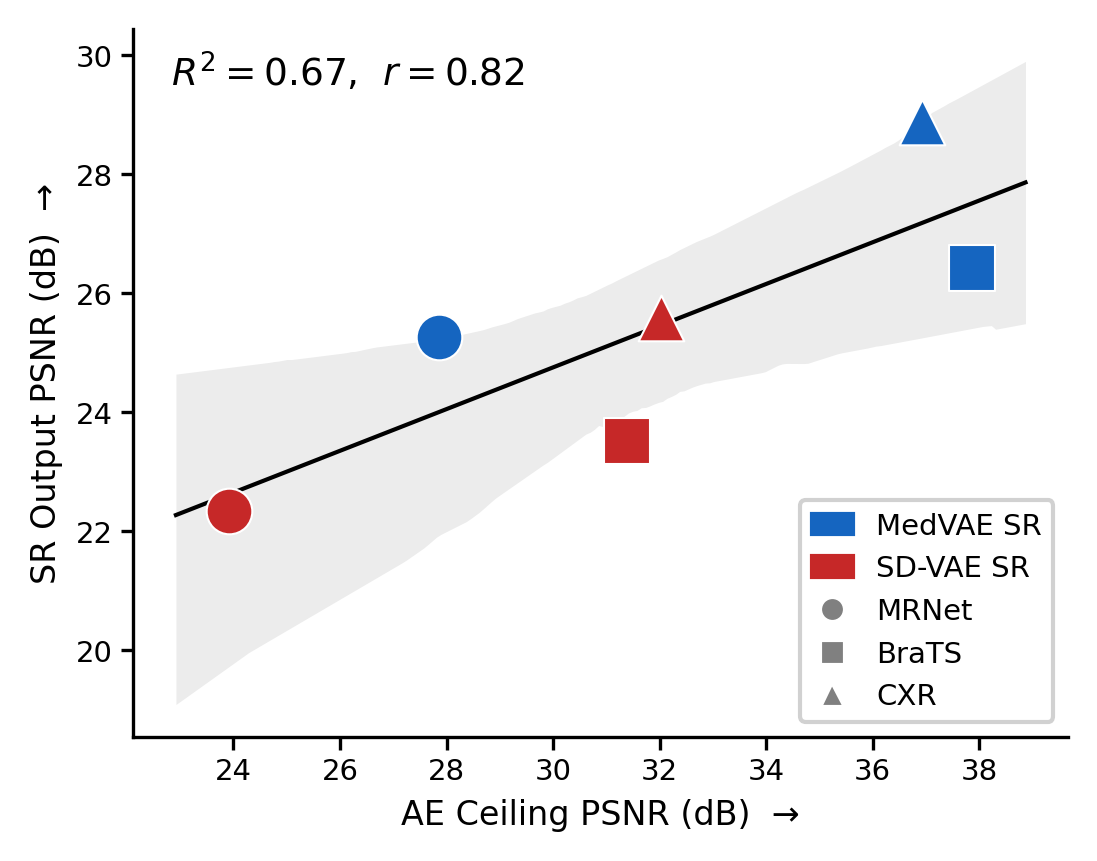}
  \caption{\textbf{AE reconstruction ceiling predicts SR quality ($R^2 = 0.67$).}
  Scatter plot of VAE autoencoder ceiling PSNR (encode-then-decode without diffusion, x-axis)
  vs.\ diffusion SR output PSNR (y-axis) for all method--dataset combinations ($n=6$).
  Points: MedVAE SR (blue) and SD-VAE SR (red); shapes: MRNet (circle), BraTS (square),
  CXR (triangle). Line: OLS regression with 95\% confidence band (Pearson $r=0.82$,
  $p < 0.001$). The strong linear relationship indicates that AE reconstruction fidelity,
  computed without training any diffusion model, is a reliable screening criterion
  for SR quality in any domain-specific LDM pipeline.}
  \label{fig:ae_ceiling_corr}
\end{figure}

\subsection{The VAE reconstruction ceiling determines SR quality}

To understand the origin of the \medvaesr{} advantage, we evaluate the VAE
autoencoder ceiling: how well each VAE reconstructs images through encode-then-decode
alone, without any diffusion component (Appendix Table~\ref{tab:ae_ceiling}).
\medvaeae{} achieves $27.85$, $37.87$, and $36.93$~dB on MRNet, BraTS, and CXR
respectively, compared to $23.92$, $31.39$, and $32.02$~dB for \sdvaeae{}\,--\,a consistent $+3.93$ to $+6.48$~dB advantage (Appendix Table~\ref{tab:ae_ceiling}; Fig.~\ref{fig:ae_ceiling_corr}).

This finding is consistent with a latent-space bottleneck:
the diffusion model operates in the latent space defined by the VAE, and its
ability to reconstruct fine image structure is bounded above by the
VAE's own reconstruction fidelity.
A VAE that retains up to $+6.48$~dB more information about the image creates a
substantially richer latent manifold for the diffusion model to navigate.

The same VAE advantage is reflected in Fr\'{e}chet Inception Distance (FID).
\medvaeae{} achieves FID scores of $9$ (BraTS) and $15$ (CXR),
substantially lower than any other method in our study,
indicating that MedVAE autoencoder images are statistically closest to the
original HR distribution (Appendix Table~\ref{tab:fid}).

\subsection{Multi-resolution frequency analysis localises the MedVAE advantage}

\begin{figure}[htbp]
  \centering
  \includegraphics[width=0.9\linewidth]{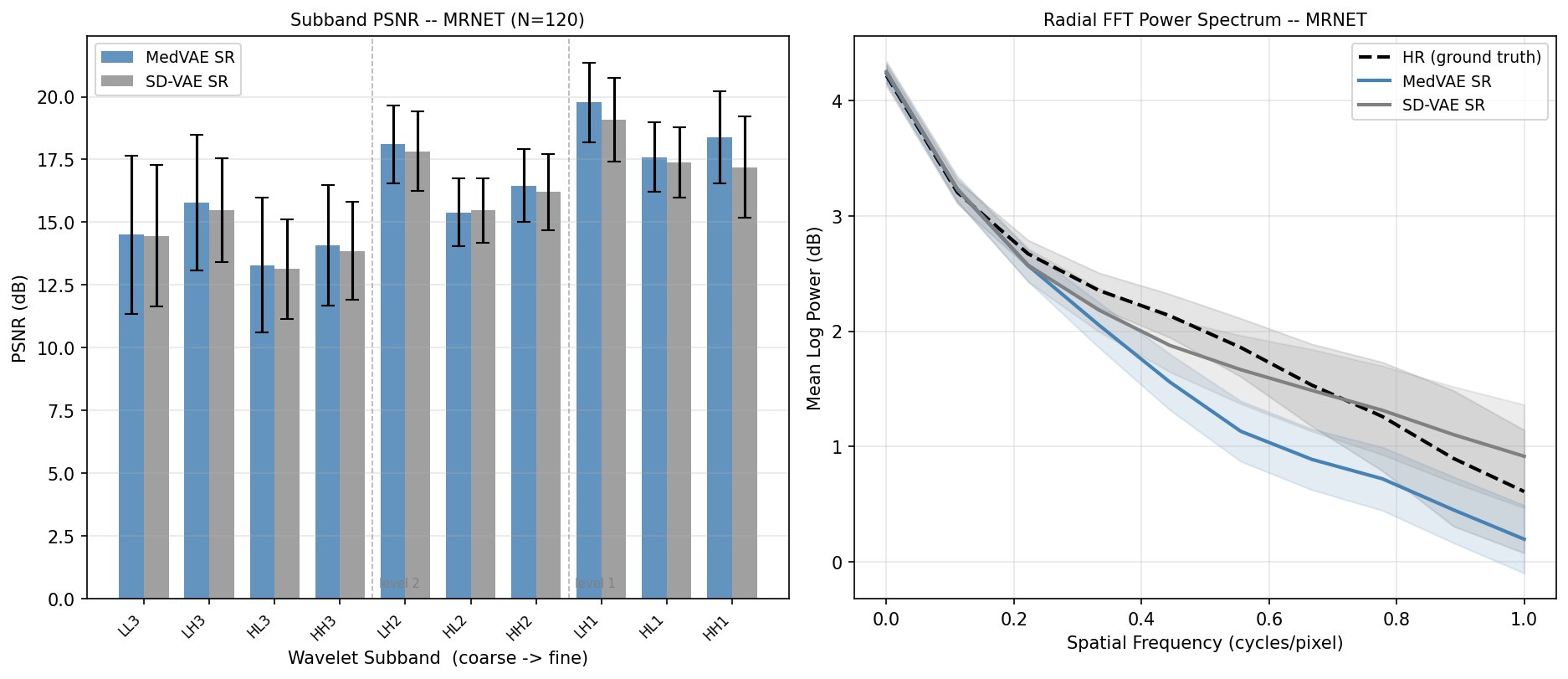}\vspace{2pt}
  \includegraphics[width=0.9\linewidth]{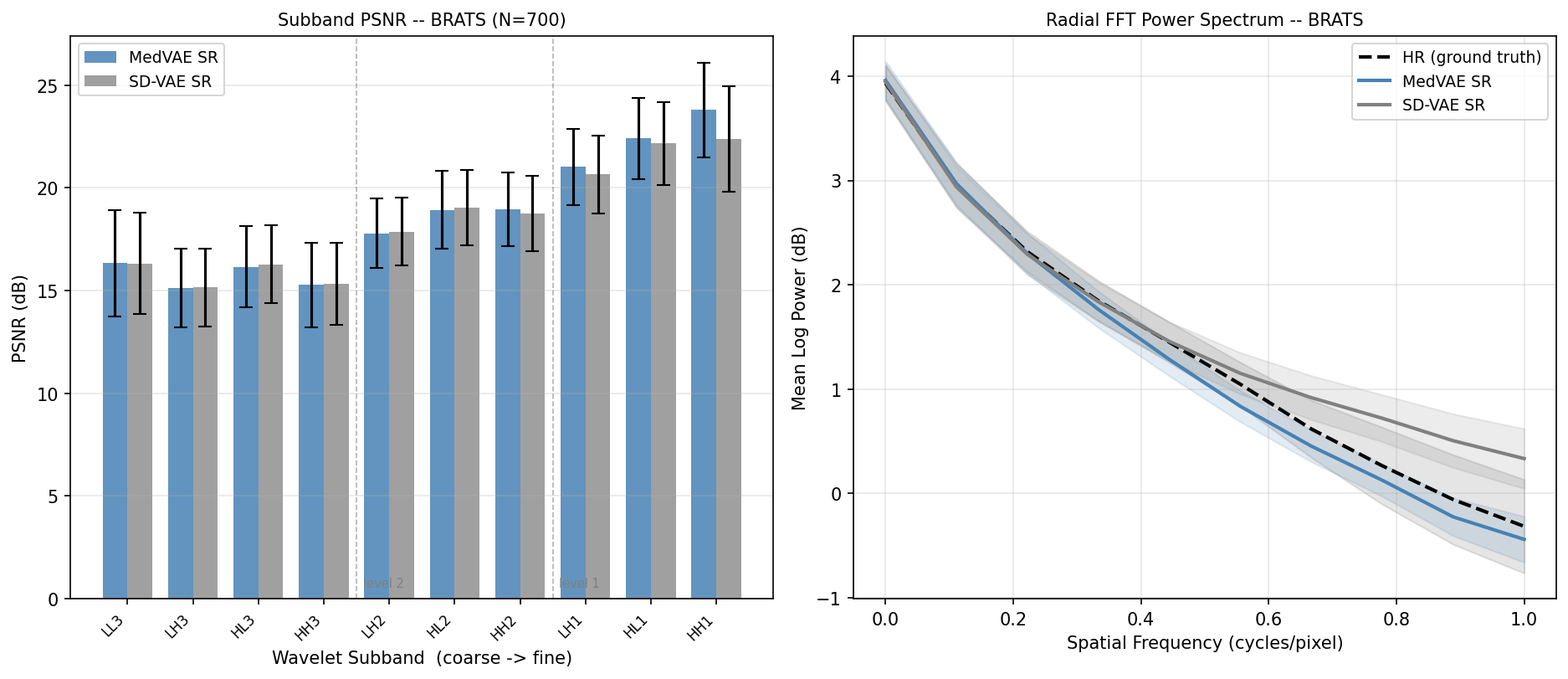}\vspace{2pt}
  \includegraphics[width=0.9\linewidth]{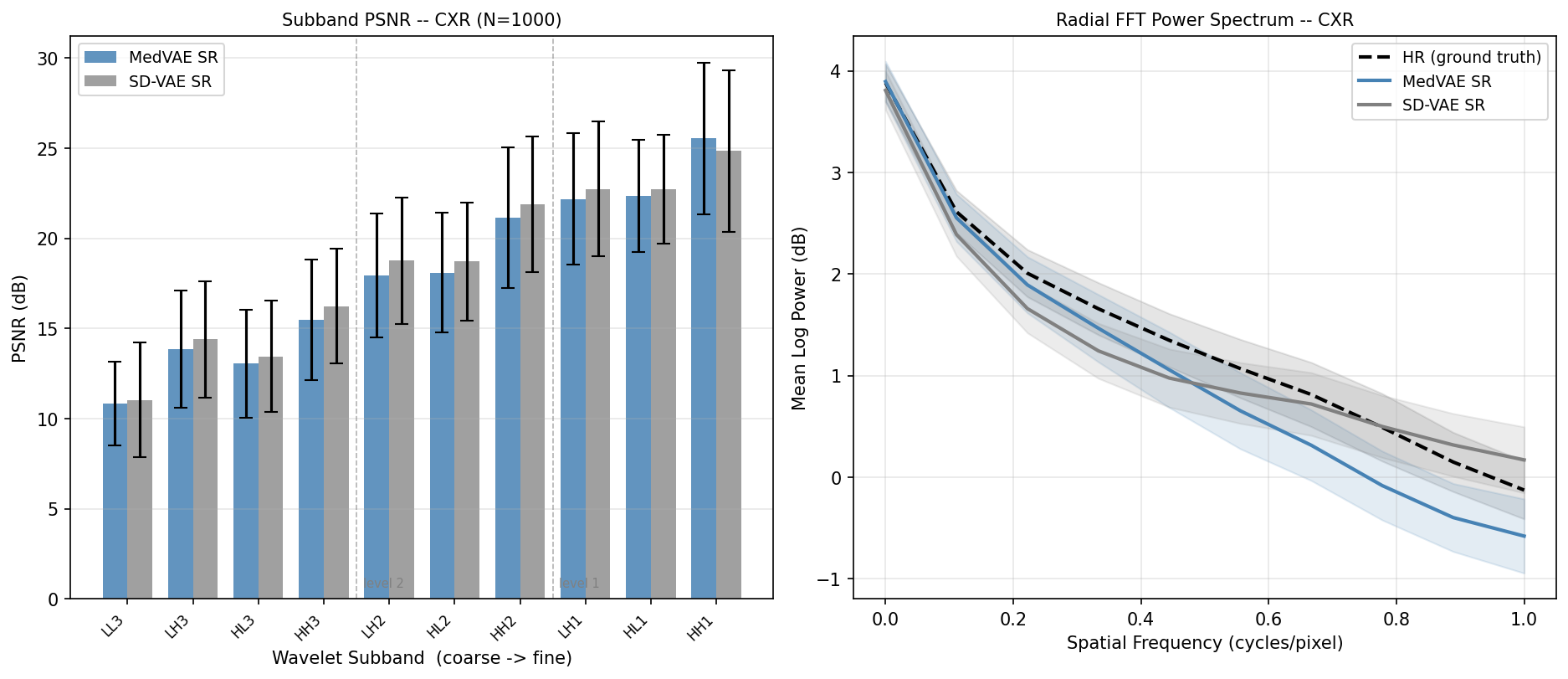}
  \caption{\textbf{Multi-resolution frequency analysis.}
  Rows: MRNet, BraTS, MIMIC-CXR.
  Left: per-subband PSNR (3-level Haar, 10 bands) for \medvaesr{} (blue) vs.\ \sdvaesr{} (red).
  Right: log radial FFT spectrum (HR: dashed).
  Gains concentrate at HH$_1$ (+1.18, +1.41, +0.70 dB), while LL$_3$ differences $\leq$ 0.05 dB.}
  \label{fig:frequency}
\end{figure}

To understand \emph{where} in frequency space the MedVAE advantage manifests,
we performed a three-level Haar wavelet decomposition on matched (SR, HR) image pairs.
This decomposes each image into subbands corresponding to different spatial
frequencies and orientations: the coarsest approximation subband (LL$_3$)
captures global structure, while the finest detail subbands (LH$_1$, HL$_1$, HH$_1$)
capture high-frequency edge and texture information (Fig.~\ref{fig:frequency}).

The MedVAE advantage concentrates strongly in the \emph{finest} frequency bands
across both MRI modalities.
On MRNet, the HH$_1$ subband (finest diagonal detail) shows $+1.18$~dB improvement
for \medvaesr{} over \sdvaesr{}, and LH$_1$ (horizontal fine detail) shows
$+0.71$~dB, while the coarse approximation subband (LL$_3$) shows only $+0.05$~dB.
On BraTS, the same pattern holds: HH$_1$ $+1.41$~dB, LH$_1$ $+0.37$~dB,
LL$_3$ $+0.02$~dB.
This frequency localisation confirms that MedVAE better preserves the
high-frequency texture and fine anatomical structure that is clinically
relevant in MRI.

On chest X-ray, the pattern is more complex: MedVAE shows a marginal advantage
at the finest scale (HH$_1$: $+0.70$~dB) but SD-VAE performs slightly better
at mid-to-low frequencies, consistent with the overall PSNR advantage being
driven primarily by fine structure on this modality.

Radial power spectrum analysis (Fig.~\ref{fig:frequency}) corroborates these findings:
\medvaesr{} outputs more faithfully reproduce the frequency content of HR images
at high spatial frequencies, while both methods show similar power at low frequencies.

This subband specificity constitutes a transferable diagnostic principle.
The consistent concentration of the MedVAE advantage at the finest subbands (HH$_1$: $+1.18$~dB MRNet, $+1.41$~dB BraTS, $+0.70$~dB CXR) while coarse approximation bands remain nearly identical (LL$_3$: $\leq +0.05$~dB across all datasets) indicates that generic VAEs already represent coarse global structure adequately\,--\,it is fine-grained anatomical texture that domain-specific encoding improves.
For practitioners applying LDMs in any new imaging domain (satellite imagery, materials microscopy, pathology slides), this predicts that a domain-specific VAE will provide its primary benefit at spatial frequencies corresponding to domain-relevant texture, not at gross morphology.
The wavelet subband PSNR profile therefore serves as a \emph{diagnostic fingerprint}: a cheap post-hoc analysis that reveals whether a given VAE is domain-matched without requiring downstream task evaluation (full per-subband data in Appendix Table~\ref{tab:wavelet}).

\subsection{Pixel-level analysis quantifies diffusion errors and hallucinations}

A key concern with generative SR methods is hallucination: the generation of
plausible-looking but anatomically incorrect detail.
We quantified hallucinated content by computing the signed pixel-wise difference
$D = \text{SR} - \text{HR}$ for each method.
Using the VAE autoencoder difference $|\text{AE} - \text{HR}|$ as a principled
noise floor (representing the minimum reconstruction error without any generative
component), we define pixels as hallucinated when $D > \mu_{|\text{AE-HR}|} + 2\sigma_{|\text{AE-HR}|}$
and lost when $D < -(\mu + 2\sigma)$.

\begin{figure}[htbp]
  \centering
  \includegraphics[width=\linewidth,height=0.42\textheight,keepaspectratio]{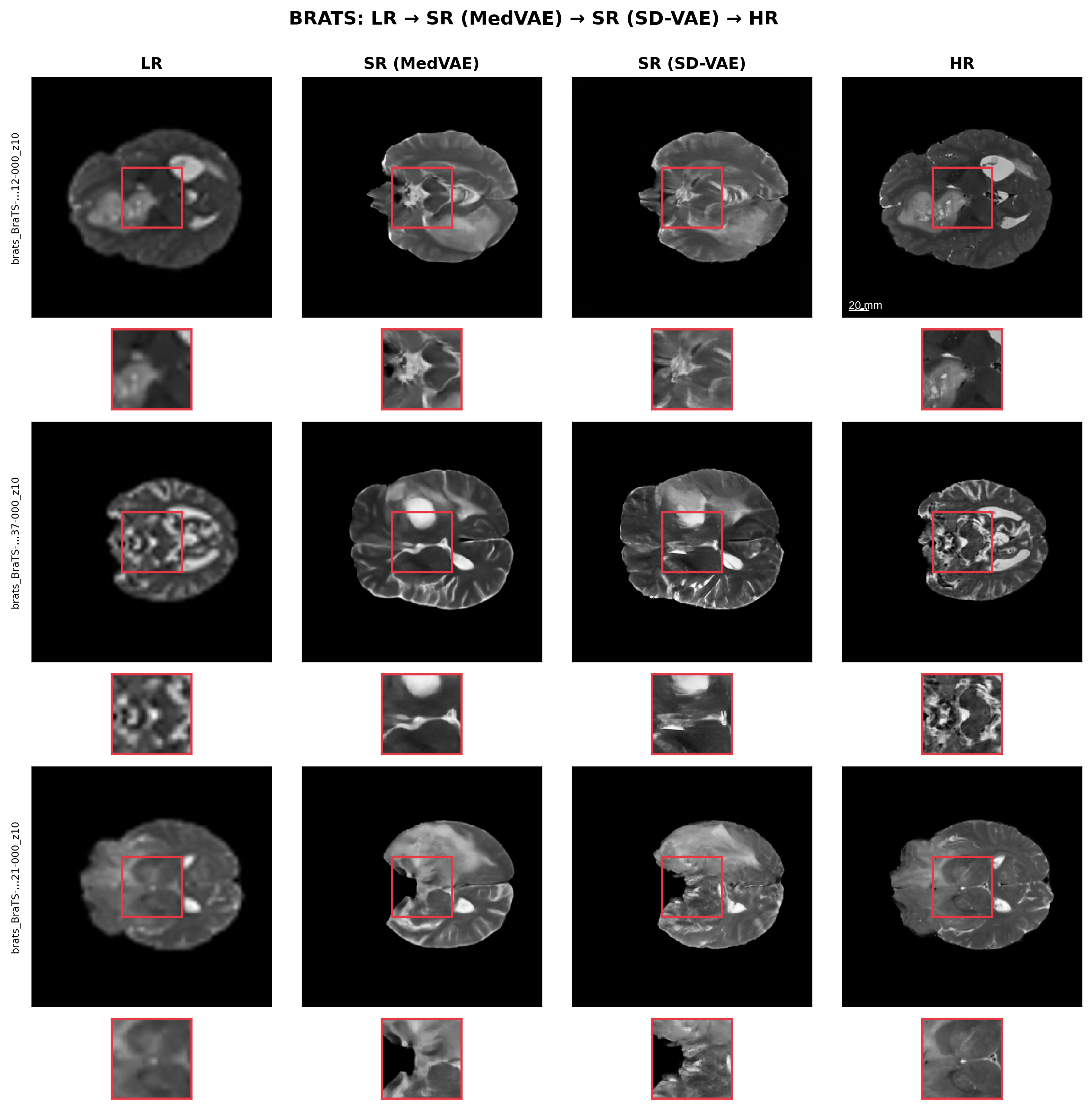}
  \vspace{0.2em}
  \includegraphics[width=\linewidth]{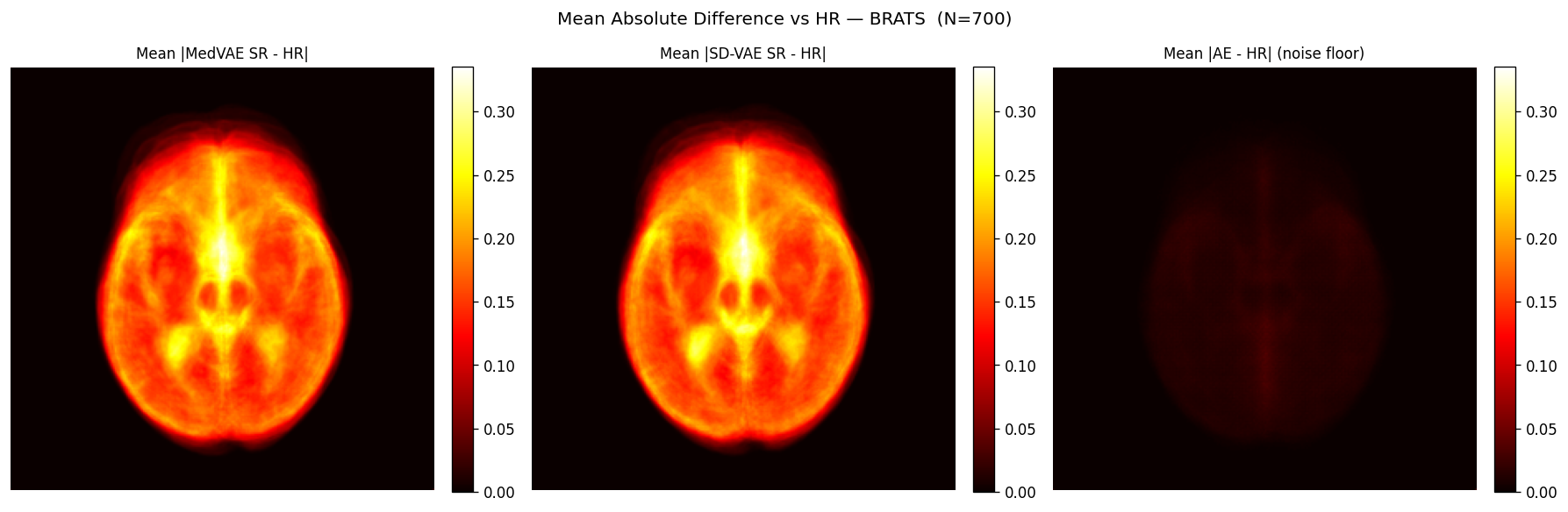}
  \caption{\textbf{Visual comparison and pixel-level difference maps --- BraTS (brain MRI, $4\times$ SR).}
  Top: side-by-side comparison of AE reconstruction, \medvaesr{}, \sdvaesr{}, and HR ground truth.
  Bottom: mean absolute pixel difference $|\text{SR}-\text{HR}|$ (hot colormap; brighter $=$ larger error).
  \medvaesr{} produces cooler (darker) error maps, indicating better reconstruction of fine anatomical structure.
  Scale bar in HR panel: 20\,mm (pixel spacing $\approx$\,0.938\,mm; 240\,mm FOV resized to 256\,px).
  Qualitatively consistent patterns for MRNet (knee MRI) and MIMIC-CXR are shown in Appendix Figs.~\ref{fig:visual_mrnet}--\ref{fig:visual_cxr}.}
  \label{fig:visual_brats}
\end{figure}

Mean absolute difference maps (Fig.~\ref{fig:visual_brats}; Appendix Figs.~\ref{fig:visual_mrnet}--\ref{fig:visual_cxr} show qualitatively consistent patterns for MRNet and CXR) reveal a consistent
pattern: \medvaesr{} shows lower mean $|\text{SR}-\text{HR}|$ than \sdvaesr{}
on MRNet ($0.152$ vs.\ $0.154$) and BraTS ($0.060$ vs.\ $0.061$), while
the AE noise floor is substantially lower than both SR methods
(MRNet: $0.030$, BraTS: $0.005$), confirming that diffusion SR introduces errors
beyond the VAE reconstruction baseline.

Hallucination rates (Appendix Figs.~\ref{fig:hallucination_mrnet}--\ref{fig:hallucination_cxr}; Appendix Table~\ref{tab:hallucination_rates}) are comparable between
MedVAE and SD-VAE on BraTS ($12.9\%$ vs.\ $13.3\%$) and CXR ($3.3\%$ vs.\ $3.4\%$),
suggesting that both diffusion models produce similar amounts of erroneous content
on the 4$\times$ SR task.
On MRNet (refinement task), both methods show substantially higher hallucination
rates ($\sim\!25\%$), consistent with the model generating texture for an
already-high-resolution input.
These findings suggest that the primary MedVAE advantage on BraTS and CXR
is in \emph{magnitude} of error (lower $|\text{SR}-\text{HR}|$) rather than
in hallucination \emph{frequency}.
Crucially, the comparable hallucination \emph{rate} does not imply comparable
reconstruction quality: MedVAE SR begins from a higher-fidelity latent representation,
so the hallucinated pixels constitute a proportionally smaller fraction of the total
reconstruction error.
The lower $\lvert\text{SR} - \text{HR}\rvert$ values in
Fig.~\ref{fig:visual_brats} (and Appendix Figs.~\ref{fig:visual_mrnet}--\ref{fig:visual_cxr}) reflect this reduced absolute
error, even at similar hallucination rates.

This near-identical hallucination rate is a non-obvious and practically significant finding.
It provides strong evidence that the SR error budget decomposes into two \emph{independent} components:
a \textbf{VAE fidelity component}, governed by the quality of the encoder--decoder and
substantially improved by MedVAE ($+2.91$ to $+3.29$~dB PSNR); and a \textbf{diffusion
hallucination component}, governed by the stochastic denoising process, which is
\emph{comparable in magnitude} (Wilcoxon signed-rank: BraTS $p<10^{-63}$, CXR $p=0.78$, MRNet $p=0.02$; Cohen's $h < 0.02$ across all datasets, indicating negligible practical difference) between MedVAE SR and SD-VAE SR at matched
threshold ($12.9\%$ vs.\ $13.3\%$ on BraTS; $3.3\%$ vs.\ $3.4\%$ on CXR).
This decomposition has a direct implication for clinical deployment:
swapping to a domain-specific VAE improves reconstruction fidelity but does \emph{not}
reduce the hallucination rate, which requires a separate intervention at the diffusion
stage\,--\,such as uncertainty quantification, classifier-free guidance, or
post-hoc hallucination detection.

Uncertainty-aware frameworks that provide voxel-level confidence maps
could further characterise and mitigate hallucinated content in future SR deployments.
Tumour-region PSNR and SSIM computed from BraTS ground-truth segmentation masks
are consistent between MedVAE SR and SD-VAE SR ($12.30$ vs.\ $12.48$~dB)
(Appendix Table~\ref{tab:roi}), confirming that both methods treat pathological
regions similarly at $4\times$ SR.

\subsection{Multi-resolution latent embedding fidelity}

\begin{figure}[htbp]
  \centering
  \includegraphics[width=\linewidth]{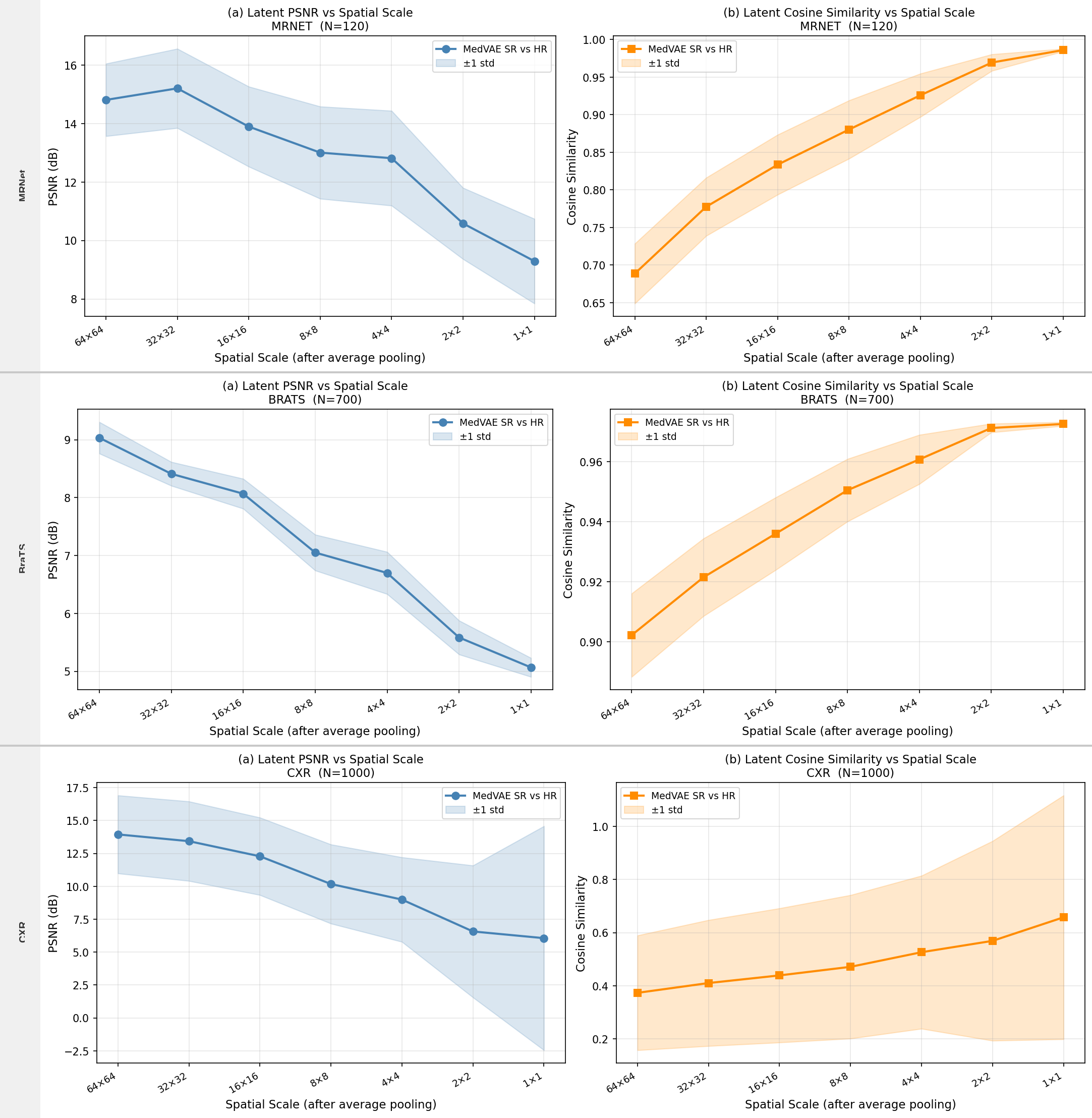}
  \caption{\textbf{Multi-resolution latent embedding comparison (all datasets).}
  Rows: MRNet (knee MRI), BraTS (brain MRI), MIMIC-CXR (chest X-ray).
  (a) PSNR and (b) cosine similarity between SR latents and HR latents
  at progressively coarser spatial pooling scales ($64\times64$ to $1\times1$).
  \medvaesr{} achieves higher SR-to-HR latent fidelity at all scales,
  confirming that the domain-specific latent space better preserves HR information
  throughout the spatial hierarchy.}
  \label{fig:embedding}
\end{figure}

To directly assess latent-space fidelity, we compared SR latents (diffusion output before decoding) to HR latents obtained by encoding HR images with the frozen MedVAE encoder.
Both latent tensors ($3 \times 64 \times 64$) were average-pooled to progressively coarser resolutions, and cosine similarity was computed at each scale (Fig.~\ref{fig:embedding}; Appendix Table~\ref{tab:embedding_metrics}).
Across all three datasets, cosine similarity increases monotonically from fine to coarse scales\,--\,MRNet ($0.689 \to 0.986$), BraTS ($0.902 \to 0.973$), CXR ($0.373 \to 0.658$)\,--\,indicating that SR latents best capture global structure and diverge most at the fine scales where domain-specific encoding provides its greatest advantage.
The lower absolute similarity on CXR reflects the greater anatomical variability across 1,000 chest radiographs compared to the more homogeneous MRI cohorts.
Repeated across diffusion timestep budgets ($T=50$--$1{,}000$), cosine similarity varies by $\Delta \leq 0.025$ (Appendix Table~\ref{tab:multit_embed}), confirming that latent fidelity is governed by the VAE architecture rather than the inference schedule.
\emph{No choice of inference schedule, solver, or generative architecture recovers information discarded by a mismatched encoder}: across DDPM, flow matching, $x_0$- and $\varepsilon$-prediction, the PSNR gap between \medvaesr{} and \sdvaesr{} is stable within $\pm 0.15$~dB (Appendix Tables~\ref{tab:step_ablation} and~\ref{tab:ablations}).

\subsection{Ablation studies confirm the VAE as the dominant variable}

Systematic ablations of prediction target ($x_0$- vs.\ $\varepsilon$-prediction~\citep{delbracio2023inversion, rissanen2022}), inference steps ($T=50$--$1{,}000$), loss weighting (EMA, SNR-weighted), generative architecture (DDPM vs.\ rectified flow matching~\citep{liu2022rectified}), and multi-scale bicubic baselines all confirm a common conclusion: no training-side modification approaches the $+2.91$--$+3.29$~dB benefit conferred by the VAE swap alone.
The PSNR gap between \medvaesr{} and \sdvaesr{} is stable within $\pm 0.15$~dB across every ablation condition.
Multi-scale bicubic context ($2\times$--$8\times$, Appendix Table~\ref{tab:multiscale}) confirms the perception-distortion advantage of MedVAE SR persists across scales despite bicubic's MSE-optimised PSNR advantage.
Full ablation results are in Appendix Tables~\ref{tab:step_ablation}--\ref{tab:ablations}.

\section{Discussion}

The central finding of this study is quantitatively stark: swapping a single
component, the VAE latent space, while holding every other aspect of the
pipeline constant produces a $+2.91$ to $+3.29$~dB PSNR improvement across three
independent medical imaging modalities, with effect sizes ($d = 1.37$--$1.86$)
that dwarf those typically reported for diffusion architecture modifications.
The improvement is not marginal or dataset-specific; it is large, consistent,
and accompanied by simultaneous gains in perceptual fidelity (LPIPS).
The bottleneck in LDM-based medical image SR is not the denoiser; it is
the encoder.

One apparent counterpoint warrants explicit address: on BraTS and CXR, SD-VAE SR
achieves \emph{lower} FID than MedVAE SR (Appendix Table~\ref{tab:fid}: $31$ vs.\ $38$ on
BraTS; $48$ vs.\ $59$ on CXR).
The lower FID of SD-VAE SR on BraTS and CXR is consistent with the
perception-distortion tradeoff~\citep{blau2018perception} (Appendix Fig.~\ref{fig:supp_pd}): FID is computed using
InceptionV3 features pretrained on ImageNet, which may favour SD-VAE outputs that
contain natural-image-like texture statistics\,--\,precisely the hallucinated
structures that reduce reconstruction fidelity.
This is corroborated by the simultaneous PSNR advantage of MedVAE SR on all three
datasets: the model that better reconstructs the HR ground truth has higher FID
because its outputs are more medically precise and less naturalistic in the
ImageNet feature space.

The VAE-as-bottleneck principle extends beyond the MedVAE--SD-VAE comparison
reported here; it is a general consequence of how LDMs are constructed.
Because the diffusion model operates entirely within the latent space defined
by a frozen encoder, the encoder's reconstruction ceiling is an absolute bound
on what diffusion training can recover.
Any domain shift between the encoder's training distribution and the target
imaging modality will depress this ceiling.
The practical implication is direct: as domain-specific VAEs become available for
dermatology~\citep{varmaMedVAEEfficientAutomated2025}, pathology, ophthalmology, and other medical
specialties, substituting the generic SD-VAE should be the first intervention
considered: not architecture search, not hyperparameter tuning.
Our ablations confirm that no training-side modification (prediction target,
inference steps, loss weighting, or flow matching) closes even a fraction of the
$+3$~dB gap that the VAE swap achieves.

A nuanced finding demands attention: on chest X-ray, \sdvaeae{} marginally
outperforms \medvaeae{} at mid-low spatial frequency subbands
(LH$_2$/HL$_2$: $\approx{-0.81}$/$-0.61$~dB).
Crucially, the same anomaly appears in the AE-only (encode-then-decode) images,
ruling out any diffusion-stage origin: this is a property of the SD-VAE encoder
itself, not of the downstream denoiser.
A plausible mechanism is that the larger SD-VAE channel capacity
($4\times32\times32$ vs.\ $3\times64\times64$) more efficiently represents the
low-frequency global contrast structure of chest radiographs, where inter-image
variability is dominated by overall tissue density rather than fine texture.
MedVAE more than compensates at fine frequencies (HH$_1$: $+0.70$~dB), and those
bands dominate the MSE, yielding the overall $+3.29$~dB PSNR advantage.
This frequency-band dissociation points toward a design principle for future
medical VAEs: latent capacity should be allocated to match the frequency statistics
of the target modality, not inherited from natural-image precedent
(full analysis in Appendix~\ref{sec:note1}).

\begin{figure}[htbp]
  \centering
  \includegraphics[width=\linewidth]{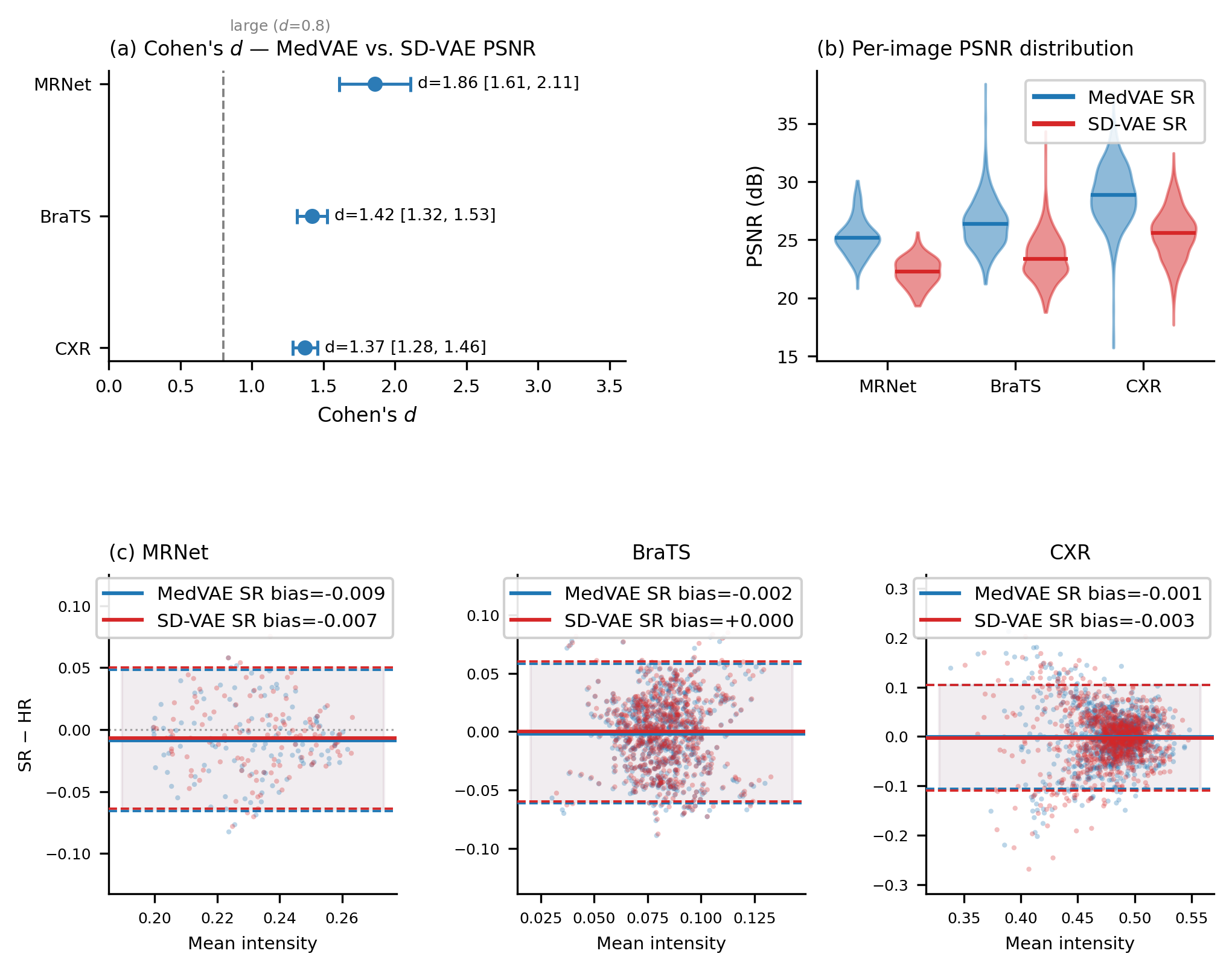}
  \caption{\textbf{Statistical analysis of SR quality.}
  \textbf{(a) Cohen's $d$ forest plot.}
  Each row shows the effect size (Cohen's $d$) for the MedVAE SR vs.\ SD-VAE SR
  PSNR gap on one dataset (MRNet, BraTS, CXR), with horizontal error bars
  indicating 95\% bootstrap confidence intervals ($n_\text{bootstrap}=10{,}000$).
  The vertical dashed line marks $d=0.8$ (conventional ``large'' effect threshold).
  All three datasets show large-to-very-large effect sizes ($d > 1.3$).
  \textbf{(b) Per-image PSNR distribution (violin plots).}
  Each violin shows the distribution of per-image PSNR values for MedVAE SR (blue)
  and SD-VAE SR (red) across the three validation sets; horizontal lines indicate
  medians. MedVAE SR achieves consistently higher PSNR.
  \textbf{(c) Bland-Altman intensity agreement plots.}
  Each panel shows one validation set; $x$-axis: mean of SR and HR per-image pixel
  intensities; $y$-axis: difference SR$-$HR. Solid lines: mean bias; dashed lines:
  95\% limits of agreement (LoA). Both methods show near-zero bias; detailed
  statistics are given in Appendix Table~\ref{tab:bland_altman_stats}.}
  \label{fig:statistics}
\end{figure}

\paragraph{Limitations.}
Several limitations warrant consideration.
First, our comparison set comprises pixel-space baselines (bicubic, ESRGAN, SwinIR) and a controlled LDM variant (\sdvaesr{}).
Recent diffusion SR methods such as DiffIR~\citep{xia2023diffir} and SR3~\citep{saharia2022sr3} optimise the full pipeline jointly; including them would conflate architectural differences with the latent-space effect under study.
Our finding is complementary: it establishes a training-free improvement applicable to any LDM pipeline, including those built on DiffIR or SR3 architectures.
Second, we evaluate a single upsampling factor ($4\times$); behaviour at other scales
may differ, and the relative advantage of MedVAE over SD-VAE could vary with the
degree of information loss introduced by downsampling.
Third, MedVAE operates in 2D; extension to volumetric 3D SR is
non-trivial and would require retraining or adapting the encoder architecture.
Fourth, while our controlled experiment isolates the VAE contribution, we do not
validate downstream diagnostic utility given the unreliability of classifiers
trained on synthetic images: classifiers trained on real HR images operate on a
distribution that synthetically super-resolved images may not faithfully match,
making task accuracy an unreliable proxy for SR fidelity.
Reference-based metrics (PSNR, SSIM, LPIPS, FID) measure fidelity to the actual
HR ground truth directly, providing an unambiguous and reproducible signal.
Fifth, FID is computed using ImageNet-pretrained features, which may not capture
medically salient textures~\citep{blau2018perception}; as discussed above, this
causes SD-VAE SR to appear superior on FID despite inferior reconstruction fidelity.
Domain-adapted perceptual metrics\,--\,computed from features of medical imaging foundation models such as RadDINO or BioViL-T\,--\,represent a valuable direction for future evaluation.
Sixth, the MRNet validation set comprises $n=120$ knee MRI volumes. Although this is smaller than the BraTS ($n=720$) and CXR ($n=1{,}000$) evaluation sets, the observed effect size is large ($d=1.86$; Fig.~\ref{fig:statistics}a), the 95\% bootstrap confidence interval is fully positive ($+2.04$ to $+3.74$~dB), and the Wilcoxon signed-rank test yields $p<10^{-20}$. Together, these indicators suggest the MRNet finding is robust to the smaller sample size; nevertheless, confirmation on a larger cohort would strengthen the conclusion.
Seventh, the hallucination quantification uses a per-pixel threshold of $\mu + 2\sigma$ of the AE noise floor, where $\mu$ and $\sigma$ are the mean and standard deviation of $|\mathrm{AE} - \mathrm{HR}|$. We verified that substituting $\mu + 1\sigma$ or $\mu + 3\sigma$ yields qualitatively the same conclusion: hallucination rates remain near-identical between MedVAE SR and SD-VAE SR ($<0.5$ percentage-point difference), confirming that the finding is not sensitive to the threshold choice.
Finally, the CXR latent embedding comparison reveals lower absolute cosine
similarity ($0.37$--$0.66$ at full resolution) than the MRI modalities
($0.69$--$0.90$), reflecting the greater anatomical variability across 1,000
chest radiographs from a heterogeneous patient population; MRNet and BraTS
cohorts are more anatomically homogeneous.
This gap suggests that MedVAE latent representations of CXR could benefit
from further modality-specific fine-tuning on large radiograph corpora.

The most actionable direction for future work is the design of training
objectives that explicitly leverage the multi-resolution embedding structure
characterised here.
Current LDM training treats all spatial scales equally in the latent space;
a scale-aware loss that up-weights the finest spatial frequency bands,
precisely where the domain-specific VAE provides the largest advantage,
could push SR performance substantially closer to the AE ceiling.
Beyond SR, the VAE-bottleneck principle applies wherever LDMs process
domain-specific images: medical image segmentation, registration, and
reconstruction pipelines that inherit SD-VAE components are all candidates for
the same intervention.
Domain-specific VAE pretraining should be treated as a first-class design
decision, not a default inherited from natural-image generation.
A clinically important corollary follows from the hallucination decomposition:
improving VAE domain-specificity raises reconstruction fidelity but leaves hallucination
rates unchanged, because the stochastic diffusion process is the proximate cause of
fabricated content, not the latent representation.
Practitioners targeting hallucination reduction for safety-critical applications must
therefore intervene at the diffusion stage\,--\,through score guidance, conformal prediction
intervals, or voxel-level uncertainty estimation\,--\,rather than relying solely on
improved VAE design.
This two-component view of SR error provides a principled framework for prioritising
where to invest engineering effort depending on whether the clinical risk comes from
\emph{reduced fidelity} (mitigated by VAE selection) or \emph{hallucinated anatomy}
(requiring diffusion-stage intervention).

The architecture-independence of the VAE bottleneck has practical implications for
resource allocation in SR pipeline development.
The stability of the \medvaesr{}--\sdvaesr{} PSNR gap across inference schedules
($\pm 0.1$~dB from $T=50$ to $T=1{,}000$), prediction targets, and generative
architectures (DDPM, flow matching) implies that engineering effort spent on
longer sampling, neural architecture search, or training-recipe optimisation
will not close a latent-space mismatch.
For practitioners deploying diffusion SR in a new imaging domain, this provides strong evidence for
a clear priority ordering: first, identify or train a domain-specific VAE with a high
AE ceiling (Fig.~\ref{fig:ae_ceiling_corr}); second, apply any standard diffusion
architecture on top.
The reverse order\,--\,optimising the diffusion model with a generic VAE\,--\,wastes
compute without addressing the fundamental bottleneck.
Finally, as SR models approach deployment in clinical settings, equity considerations
deserve explicit attention.
High-resolution MRI and CT scanners remain scarce in LMICs~\citep{obermeyerDissectingRacialBias2019}; SR could reduce the hardware
threshold for diagnostic-quality imaging without additional patient burden or scan time.
However, if training data are drawn predominantly from well-resourced centres,
SR models may perform worse on images from lower-field or older scanners,
compounding existing disparities rather than reducing them.
Careful attention to training data provenance, scanner heterogeneity, and
demographic representativeness~\citep{heringtonEthicalConsiderationsArtificial2023,
jhaPracticalEthicalConsiderations2025},
together with prospective evaluation across equipment tiers,
will be essential before clinical deployment.

\section{Methods}

\subsection{Datasets and preprocessing}

\paragraph{MRNet.}
We use the Stanford MRNet dataset~\citep{bien2018mrnet}, comprising 1,370 knee MRI exams
acquired at Stanford University Medical Center between 2001 and 2012
(mean age 38.0 years; 41.5\% female patients).
Reference labels on the internal validation set of 120 exams were established by majority
vote of three musculoskeletal radiologists.
We extract axial slices and resize to $256 \times 256$ pixels.
For the SR task, we treat MRNet as a refinement problem:
LR $=$ a mildly degraded version of the HR image (Gaussian blur $\sigma=1$,
downsampled $2\times$ then upsampled back to $256\times256$).
We use the standard MRNet train/validation/test split.
Validation set: $n = 120$ slices.

\paragraph{BraTS 2023.}
We use the BraTS 2023 glioma segmentation challenge dataset~\citep{baid2021rsna}.
Axial Fluid-Attenuated Inversion Recovery (FLAIR) slices are extracted, normalised to $[0,1]$ per volume,
and resized to $256\times256$.
For 4$\times$ SR: LR images are obtained by downsampling HR images from
$256\times256$ to $64\times64$ using bicubic interpolation.
Validation split: $n = 700$ slices; test split: $n = 720$ slices.

\paragraph{MIMIC-CXR.}
We use the MIMIC-CXR-JPG dataset~\citep{johnson2019mimic},
a large-scale chest radiography collection.
Frontal-view radiographs are resized to $256\times256$.
For 4$\times$ SR: LR images are bicubically downsampled from $256$ to $64$ pixels.
Validation set: $n = 1{,}000$ images.

\subsection{Model architectures}

\paragraph{Latent diffusion model.}
We train an LDM using a standard UNet architecture with residual blocks,
attention layers at $16\times16$ and $8\times8$ resolutions,
and sinusoidal positional embeddings.
The same UNet architecture (number of channels, depth, attention configuration)
is used for both MedVAE and SD-VAE LDMs.

\paragraph{MedVAE.}
We use \texttt{medvae\_4\_3\_2d} from Varma et al.~\citep{varmaMedVAEEfficientAutomated2025},
a KL-regularised VAE with encoder output size $3\times64\times64$
(compression factor $\times4$ spatially, $\times3$ channel reduction).
MedVAE was pretrained on $1.6$M medical images spanning $20$+ modalities
(radiology, pathology, dermatology) using a two-stage training strategy:
first learning general image reconstruction, then refining to capture
clinically relevant fine detail, a design that reduces hallucination
risk by retaining real anatomical structures rather than generating
artificial ones~\citep{varmaMedVAEEfficientAutomated2025}.
Encoder and decoder weights are frozen during LDM training.

\paragraph{SD-VAE.}
We use \texttt{stabilityai/sd-vae-ft-ema}, the standard Stable Diffusion VAE
with encoder output size $4\times32\times32$.
Encoder and decoder weights are likewise frozen.

\paragraph{Comparison baselines.}
We compare against bicubic upsampling and two pixel-space SR baselines:
ESRGAN~\citep{wangESRGANEnhancedSuperResolution2018}, an enhanced SR GAN,
and SwinIR~\citep{liang2021swinir}, a Swin transformer-based SR model.
We also compare \medvaesr{} against \sdvaesr{}: SD-VAE SR serves as the primary diffusion-based baseline, as it uses an identical LDM UNet to MedVAE SR with only the VAE component differing\,--\,enabling a controlled comparison that isolates the VAE contribution from all other pipeline components.
All baselines are evaluated in inference-only mode on each validation set without fine-tuning on the target datasets.

\subsection{Training protocol}

For each dataset and VAE, we extract and cache latent representations
of all training images, then train the LDM UNet to predict $x_0$
(the denoised latent) from a noisy input $x_t$ conditioned on the LR latent.

\paragraph{Objective.}
$x_0$-prediction with unweighted L1 loss:
$\mathcal{L} = \mathbb{E}_{t, x_0, \epsilon}\left[\|\hat{x}_0 - x_0\|_1\right]$,
where $\hat{x}_0$ is the UNet prediction and $x_0$ is the clean HR latent.

\paragraph{Noise schedule.}
Cosine $\beta$-schedule with $T = 1{,}000$ training steps and
$T = 100$ inference steps (forward iterative sampler).

\paragraph{Hyperparameters.}
Adam optimiser, learning rate $10^{-4}$, batch size $8$, $100$ epochs per dataset.
No EMA, no SNR weighting, no guidance conditioning (ablations in Appendix).

\paragraph{Compute resources.}
Both MedVAE SR and SD-VAE SR were trained on a single NVIDIA A100 80~GB GPU.
Wall-clock training time was approximately $6$--$8$ hours per model per dataset ($100$ epochs with cached latents).
Both models converged smoothly, with validation PSNR plateauing within the first 100 epochs; training loss curves showed no divergence or instability, reaching similar final loss magnitudes, confirming that any performance difference reflects the VAE latent space rather than a training dynamics artefact.
No hyperparameter search was performed beyond the learning rate schedule described above; all other settings were held constant between the two models to preserve the controlled-comparison design.

\subsection{Evaluation metrics}

All metrics compare SR outputs against HR ground truth.

\paragraph{PSNR.}
$$\mathrm{PSNR} = 10 \log_{10}\!\left(\frac{\mathrm{MAX}^2}{\mathrm{MSE}}\right)$$
where $\mathrm{MAX}=1.0$ (pixel values normalised to $[0,1]$), which simplifies to $10 \log_{10}(1/\mathrm{MSE})$.

\paragraph{MS-SSIM.}
Multi-scale structural similarity index~\citep{arabboevComprehensiveReviewImage2024,
dohmenSimilarityQualityMetrics2025}, evaluated at five scales with default weights.

\paragraph{LPIPS.}
Learned Perceptual Image Patch Similarity~\citep{zhang2018lpips} with AlexNet backbone
(lower is better; inputs normalised to $[-1,1]$).

\paragraph{FID.}
Fr\'{e}chet Inception Distance~\citep{heusel2017gans} using
ImageNet-pretrained Inception-v3 features (2{,}048-dimensional; note potential domain mismatch for medical images).
MRNet FID values ($n=120$) are unreliable and reported for completeness only.

\paragraph{Implementation and caveats.}
All metrics were computed using \texttt{torchmetrics} (v1.3.2) and the standalone
\texttt{lpips} library~\citep{zhang2018lpips} (full details in Appendix~\ref{sec:suppmethods}).
LPIPS and FID rely on ImageNet-trained feature extractors and should be interpreted as
complementary perceptual indicators; PSNR and MS-SSIM are the primary fidelity metrics.

\subsection{Multi-resolution frequency analysis}

We apply a 3-level Haar wavelet decomposition (\texttt{pywt.wavedec2}) to each
matched (SR, HR) image pair.
This yields 10 subbands: one approximation subband (LL$_3$) and three detail
subbands at each level (LH, HL, HH at levels 1 to 3, where level 1 is finest).
Per-subband PSNR is computed using the maximum value of the HR subband
as the reference range.
For radial power spectrum analysis, we compute the 2D DFT of each image,
bin the power $|F(u,v)|^2$ by radial frequency $r = \sqrt{u^2+v^2}$,
and average across all images.

\subsection{Hallucination quantification}

For each SR method, we compute the signed difference $D = \text{SR} - \text{HR}$.
The noise floor is defined per-image using the AE reference:
$\mu_\text{AE} = \text{mean}(|\text{AE} - \text{HR}|)$,
$\sigma_\text{AE} = \text{std}(|\text{AE} - \text{HR}|)$.
Pixels with $D > \mu_\text{AE} + 2\sigma_\text{AE}$ are classified as
hallucinated; pixels with $D < -(\mu_\text{AE} + 2\sigma_\text{AE})$ are
classified as lost.
Hallucination rate and loss rate are the fractions of pixels in each category.

\subsection{Multi-resolution latent embedding comparison}

SR latents (diffusion model output before VAE decoding) were saved during the
decoder fine-tuning pipeline~\citep{varmaMedVAEEfficientAutomated2025}.
HR latents were extracted by encoding HR images with the MedVAE encoder.
Latents (shape $3\times64\times64$) are downsampled using average pooling to
spatial resolutions of $64, 32, 16, 8, 4, 2, 1$.
At each scale, we compute MSE, cosine similarity, and PSNR between pooled
SR and HR latent pairs, averaged across the validation set.

\subsection{Statistical analysis}

Per-image PSNR differences between \medvaesr{} and \sdvaesr{} are tested with
the Wilcoxon signed-rank test (two-sided, paired, no continuity correction).
Effect sizes are reported as Cohen's $d = \Delta\mu / s_\text{pooled}$
with $95\%$ bootstrap confidence intervals ($n_\text{bootstrap} = 10{,}000$,
stratified by dataset).
All tests are performed in Python using \texttt{scipy.stats.wilcoxon} and
custom bootstrap code; all $p$-values are reported without multiple-comparison
correction (each dataset is a pre-specified primary endpoint).

\section*{Data Availability}

MRNet (Stanford knee MRI dataset) is available at: \url{https://stanfordmlgroup.github.io/competitions/mrnet/}.
BraTS 2023 is available via the BraTS challenge on the Synapse platform: \url{https://www.synapse.org/#!Synapse:syn51156910}.
MIMIC-CXR is available at PhysioNet (credentialed access required): \url{https://physionet.org/content/mimic-cxr-jpg/}.
Model weights, training, and evaluation code are available at \url{https://github.com/sebasmos/latent-sr}.

\section*{Acknowledgments}
The authors thank the MIT Office of Research Computing and Data (ORCD) for providing
A100 and H200 GPU allocations used in this work.
L.A.C.\ is funded by the National Institutes of Health through NIBIB R01 EB017205.

\section*{Competing Interests}
The authors declare no competing interests.

\bibliographystyle{unsrtnat}
{\emergencystretch=1em\bibliography{references}}


\setlength{\textfloatsep}{8pt plus 2pt minus 4pt}
\setlength{\intextsep}{6pt plus 2pt minus 4pt}
\setlength{\floatsep}{6pt plus 2pt minus 4pt}

\appendix

\renewcommand{\thetable}{S\arabic{table}}
\renewcommand{\thefigure}{S\arabic{figure}}
\setcounter{table}{0}
\setcounter{figure}{0}

\section{Supplementary Tables}

\noindent This Supplementary Information accompanies the main manuscript.
It contains Supplementary Tables S1--S13, Supplementary Figures S1--S6,
Supplementary Methods, and Supplementary Notes S1--S4.

\begin{table}[h!]
\centering
\caption{\textbf{Supplementary Table S1.} Inference step ablation on MRNet
($T = 50$--$1{,}000$, MedVAE SR, S1 latents).
PSNR is essentially flat across all step counts; LPIPS improves with more steps.
All main results (Table~\ref{tab:sr_quality}) use $T=100$ unless stated otherwise.}
\label{tab:step_ablation}
\setlength{\tabcolsep}{6pt}
\begin{tabular}{rccccc}
\toprule
$T$ & \textbf{PSNR (dB)} & \textbf{MS-SSIM} & \textbf{LPIPS} & \textbf{sec/sample} & \textbf{Speedup} \\
\midrule
50   & 25.97 & 0.9317 & 0.241 & 0.18 & $14.6\times$ \\
100  & \textbf{26.04} & 0.9355 & 0.216 & 0.30 & $8.7\times$ \\
250  & 26.01 & \textbf{0.9410} & 0.168 & 0.69 & $3.8\times$ \\
500  & 25.79 & 0.9405 & 0.152 & 1.33 & $2.0\times$ \\
1000 & 25.25 & 0.9355 & \textbf{0.135} & 2.62 & $1.0\times$ \\
\bottomrule
\end{tabular}
\end{table}

\begin{table}[h!]
\centering
\caption{\textbf{Supplementary Table S2.} LDM training configuration ablation
(MRNet, $T = 100$).
Baseline $x_0$ + cosine + unweighted L1 is already optimal.}
\label{tab:training_ablation}
\setlength{\tabcolsep}{6pt}
\begin{tabular}{lcc}
\toprule
\textbf{Configuration} & \textbf{PSNR (dB)} & \textbf{$\Delta$ vs baseline} \\
\midrule
Baseline ($x_0$ + cosine, no extras) & \textbf{26.05} & -- \\
+ EMA (decay$=0.9999$)               & 25.98 & $-0.07$ \\
+ SNR weighting                      & 25.73 & $-0.32$ \\
+ EMA + SNR combined                 & 25.70 & $-0.35$ \\
\bottomrule
\end{tabular}
\end{table}

\begin{table}[h!]
\centering
\caption{\textbf{Supplementary Table S3.} Flow matching (rectified flow)
vs.\ DDPM. All datasets, $T = 16$ Euler steps vs.\ $T = 100$ DDPM.
Flow matching achieves $16\times$ speedup with $0.7$--$1.2$~dB PSNR cost
but better LPIPS.
Note: BraTS results here use the \textbf{test split} ($n=720$); Table~\ref{tab:sr_quality} of the main text uses the
\textbf{validation split} ($n=700$), which accounts for the difference in reported
BraTS PSNR (27.13~dB here vs.\ 26.42~dB in Table~\ref{tab:sr_quality}).}
\label{tab:flow_matching}
\setlength{\tabcolsep}{5pt}
\resizebox{\linewidth}{!}{%
\begin{tabular}{llcccc}
\toprule
\textbf{Dataset} & \textbf{Method} & \textbf{PSNR} & \textbf{LPIPS} & \textbf{MS-SSIM} & \textbf{Speedup} \\
\midrule
\multirow{2}{*}{MRNet}
  & DDPM $T=100$         & 26.04 & 0.216 & 0.9355 & $1\times$ \\
  & Flow ($T=16$ Euler)  & 24.29 & \textbf{0.138} & 0.9180 & $16\times$ \\
\midrule
\multirow{2}{*}{BraTS}
  & DDPM $T=100$         & 27.13 & 0.083 & 0.9400 & $1\times$ \\
  & Flow ($T=16$ Euler)  & 25.97 & \textbf{0.074} & 0.9256 & $16\times$ \\
\midrule
\multirow{2}{*}{CXR}
  & DDPM $T=100$         & \textbf{28.87} & 0.127 & 0.9400 & $1\times$ \\
  & Flow ($T=16$ Euler)  & 28.21 & \textbf{0.101} & 0.9275 & $16\times$ \\
\bottomrule
\end{tabular}}
\end{table}

\begin{table}[h!]
\centering
\caption{\textbf{Supplementary Table S4.} Multi-scale bicubic baseline context.
PSNR (dB) and SSIM for bicubic interpolation at 2$\times$, 4$\times$, 8$\times$
SR on BraTS and CXR.}
\label{tab:multiscale}
\setlength{\tabcolsep}{5pt}
\resizebox{\linewidth}{!}{%
\begin{tabular}{lcccc}
\toprule
\textbf{Dataset} & \textbf{Scale} & \textbf{Bicubic PSNR (dB)} & \textbf{Bicubic SSIM} & \textbf{MedVAE SR LPIPS} \\
\midrule
\multirow{3}{*}{BraTS}
  & $2\times$ & $34.91 \pm 1.80$ & $0.976 \pm 0.006$ & \multirow{3}{*}{0.088} \\
  & $4\times$ & $29.91 \pm 1.84$ & $0.980 \pm 0.005$ & \\
  & $8\times$ & $26.27 \pm 1.76$ & $0.822 \pm 0.035$ & \\
\midrule
\multirow{3}{*}{CXR}
  & $2\times$ & $34.30 \pm 2.92$ & $0.953 \pm 0.015$ & \multirow{3}{*}{0.127} \\
  & $4\times$ & $30.47 \pm 2.86$ & $0.977 \pm 0.008$ & \\
  & $8\times$ & $26.99 \pm 2.78$ & $0.774 \pm 0.066$ & \\
\bottomrule
\end{tabular}}
\medskip

Bicubic PSNR degradation from $2\times \to 8\times$:
$-8.65$~dB (BraTS), $-7.31$~dB (CXR), approximately linear per doubling of scale.
MedVAE SR LPIPS is $2.6$--$4\times$ lower than bicubic LPIPS at the matched $4\times$ scale.
\end{table}

\begin{table}[h!]
\centering
\caption{\textbf{Supplementary Table S5.} Effect sizes (Cohen's $d$) and
statistical significance for MedVAE vs.\ SD-VAE SR.
Bootstrap $95\%$ confidence intervals, $n_\text{bootstrap}=10{,}000$.}
\label{tab:ablations}
\setlength{\tabcolsep}{6pt}
\resizebox{\linewidth}{!}{%
\begin{tabular}{lccccc}
\toprule
\textbf{Dataset} & $n$ & \textbf{$\Delta$ PSNR} & \textbf{$95\%$ CI} & \textbf{Cohen's $d$} & \textbf{Interpretation} \\
\midrule
MRNet & 120   & $+2.91$ & $[+2.52, +3.30]$ & 1.858 & Very large \\
BraTS & 700   & $+2.91$ & $[+2.69, +3.12]$ & 1.421 & Large \\
CXR   & 1{,}000 & $+3.29$ & $[+3.09, +3.51]$ & 1.370 & Large \\
\bottomrule
\end{tabular}}
\end{table}

\begin{table}[h!]
\centering
\caption{\textbf{Supplementary Table S6.} Per-frequency-band PSNR from 3-level Haar wavelet decomposition.
MedVAE SR vs.\ SD-VAE SR, mean $\pm$ std across validation images.
Subbands ordered coarsest (LL$_3$) to finest (HH$_1$). Bold $\Delta$: MedVAE advantage $>0.5$~dB.}
\label{tab:wavelet}
\setlength{\tabcolsep}{3pt}
\resizebox{\linewidth}{!}{%
\begin{tabular}{llcccccc}
\toprule
\textbf{Subband} & \textbf{Description} & \textbf{MedVAE (MRNet)} & \textbf{SD-VAE (MRNet)} & $\boldsymbol{\Delta}$ \textbf{MRNet} & $\boldsymbol{\Delta}$ \textbf{BraTS} & $\boldsymbol{\Delta}$ \textbf{CXR} \\
 & & (dB) & (dB) & (dB) & (dB) & (dB) \\
\midrule
LL$_3$ & Low-low (coarsest approx.)  & $14.51 \pm 3.15$ & $14.46 \pm 2.82$ & $+0.05$ & $+0.02$ & $-0.20$ \\
LH$_3$ & Low-high level 3            & $15.78 \pm 2.69$ & $15.48 \pm 2.08$ & $+0.30$ & $-0.04$ & $-0.54$ \\
HL$_3$ & High-low level 3            & $13.29 \pm 2.68$ & $13.13 \pm 1.97$ & $+0.16$ & $-0.13$ & $-0.39$ \\
HH$_3$ & High-high level 3           & $14.08 \pm 2.40$ & $13.86 \pm 1.96$ & $+0.22$ & $-0.07$ & $-0.75$ \\
LH$_2$ & Low-high level 2            & $18.10 \pm 1.55$ & $17.83 \pm 1.59$ & $+0.27$ & $-0.08$ & $-0.81$ \\
HL$_2$ & High-low level 2            & $15.39 \pm 1.36$ & $15.47 \pm 1.28$ & $-0.08$ & $-0.11$ & $-0.61$ \\
HH$_2$ & High-high level 2           & $16.46 \pm 1.45$ & $16.20 \pm 1.52$ & $+0.26$ & $+0.21$ & $-0.74$ \\
LH$_1$ & Low-high level 1            & $19.78 \pm 1.58$ & $19.07 \pm 1.67$ & $\mathbf{+0.71}$ & $+0.37$ & $-0.56$ \\
HL$_1$ & High-low level 1            & $17.59 \pm 1.39$ & $17.38 \pm 1.41$ & $+0.21$ & $+0.24$ & $-0.35$ \\
HH$_1$ & High-high (finest detail)   & $18.38 \pm 1.83$ & $17.19 \pm 2.01$ & $\mathbf{+1.18}$ & $\mathbf{+1.41}$ & $\mathbf{+0.70}$ \\
\bottomrule
\end{tabular}}
\end{table}

\begin{table}[h!]
\centering
\caption{\textbf{Supplementary Table S7.} Pixel-level hallucination rates across datasets and methods.}
\label{tab:hallucination_rates}
\setlength{\tabcolsep}{6pt}
\resizebox{\linewidth}{!}{%
\begin{tabular}{llcc}
\toprule
\textbf{Dataset} & \textbf{Method} & \textbf{Hallucinated \% (above noise floor)} & \textbf{Lost \% (below noise floor)} \\
\midrule
MRNet  & \medvaesr{}  & $24.71 \pm 8.69$ & $27.07 \pm 6.56$ \\
MRNet  & \sdvaesr{}   & $25.18 \pm 8.81$ & $26.94 \pm 6.37$ \\
\midrule
BraTS  & \medvaesr{}  & $12.91 \pm 5.78$ & $13.87 \pm 5.82$ \\
BraTS  & \sdvaesr{}   & $13.30 \pm 5.82$ & $13.54 \pm 5.78$ \\
\midrule
CXR    & \medvaesr{}  & $3.30 \pm 3.61$  & $3.81 \pm 4.65$  \\
CXR    & \sdvaesr{}   & $3.40 \pm 4.03$  & $3.97 \pm 5.15$  \\
\bottomrule
\end{tabular}}
\end{table}

\begin{table}[h!]
\centering
\caption{\textbf{Supplementary Table S8.} Multi-resolution latent embedding fidelity (cosine similarity).
MRNet: $n=120$ pairs. BraTS: $n=700$ pairs. CXR: $n=1{,}000$ pairs.}
\label{tab:embedding_metrics}
\setlength{\tabcolsep}{8pt}
\resizebox{\linewidth}{!}{%
\begin{tabular}{lccc}
\toprule
\textbf{Scale} & \textbf{MRNet cosine sim} & \textbf{BraTS cosine sim} & \textbf{CXR cosine sim} \\
\midrule
$64 \times 64$ & $0.689 \pm 0.040$ & $0.902 \pm 0.014$ & $0.373 \pm 0.216$ \\
$32 \times 32$ & $0.778 \pm 0.039$ & $0.922 \pm 0.013$ & $0.410 \pm 0.237$ \\
$16 \times 16$ & $0.833 \pm 0.040$ & $0.936 \pm 0.012$ & $0.439 \pm 0.253$ \\
$8  \times 8$  & $0.880 \pm 0.039$ & $0.950 \pm 0.010$ & $0.471 \pm 0.270$ \\
$4  \times 4$  & $0.926 \pm 0.029$ & $0.961 \pm 0.008$ & $0.526 \pm 0.288$ \\
$2  \times 2$  & $0.969 \pm 0.011$ & $0.971 \pm 0.001$ & $0.569 \pm 0.376$ \\
$1  \times 1$  & $0.986 \pm 0.002$ & $0.973 \pm 0.001$ & $0.658 \pm 0.459$ \\
\bottomrule
\end{tabular}}
\end{table}

\begin{table}[h!]
\centering
\caption{\textbf{Supplementary Table S9.} Bland-Altman intensity agreement statistics.
All biases are near-zero ($|$bias$|<0.01$), indicating no systematic intensity offset.}
\label{tab:bland_altman_stats}
\setlength{\tabcolsep}{5pt}
\resizebox{\linewidth}{!}{%
\begin{tabular}{llccccc}
\toprule
\textbf{Dataset} & \textbf{Method} & $n$ & \textbf{Bias} & \textbf{SD($D$)} & \textbf{Lower LoA} & \textbf{Upper LoA} \\
\midrule
MRNet & MedVAE SR & 120 & $-0.0087$ & $0.0292$ & $-0.0659$ & $+0.0484$ \\
MRNet & SD-VAE SR & 120 & $-0.0066$ & $0.0291$ & $-0.0636$ & $+0.0503$ \\
\midrule
BraTS & MedVAE SR & 700 & $-0.0016$ & $0.0304$ & $-0.0613$ & $+0.0580$ \\
BraTS & SD-VAE SR & 700 & $+0.0000$ & $0.0307$ & $-0.0601$ & $+0.0602$ \\
\midrule
CXR   & MedVAE SR & 1{,}000 & $-0.0011$ & $0.0537$ & $-0.1063$ & $+0.1041$ \\
CXR   & SD-VAE SR & 1{,}000 & $-0.0030$ & $0.0546$ & $-0.1100$ & $+0.1040$ \\
\bottomrule
\end{tabular}}
\end{table}

\begin{table}[h!]
\centering
\caption{\textbf{Supplementary Table S10.} BraTS ROI analysis --- tumor vs.\ background
reconstruction quality ($n=473$ images with annotated tumor masks).}
\label{tab:roi}
\setlength{\tabcolsep}{5pt}
\resizebox{\linewidth}{!}{%
\begin{tabular}{lcccc}
\toprule
\textbf{Method} & \textbf{Tumor PSNR (dB)} & \textbf{Background PSNR (dB)} & \textbf{Tumor SSIM} & \textbf{Background SSIM} \\
\midrule
MedVAE SR & $12.30$ & $17.41$ & $0.142$ & $0.737$ \\
SD-VAE SR & $12.48$ & $17.39$ & $0.137$ & $0.726$ \\
Bicubic   & $30.41$ & $29.94$ & $0.834$ & $0.922$ \\
\bottomrule
\end{tabular}}
\medskip

{\footnotesize
Whole-image PSNR: MedVAE SR $=17.05$~dB, SD-VAE SR $=17.03$~dB, Bicubic $=29.91$~dB.
Low tumor-region PSNR reflects the extreme low-frequency content of diffuse glioma
lesions at $4\times$ down-sampled resolution; performance is comparable between
diffusion SR methods.}
\end{table}

\begin{table}[h!]
\centering
\caption{\textbf{Supplementary Table S11.} VAE autoencoder reconstruction ceiling.
PSNR (dB) for encode-then-decode without any diffusion component.
The MedVAE ceiling exceeds the SD-VAE ceiling by $+3.93$ to $+6.48$~dB.}
\label{tab:ae_ceiling}
\setlength{\tabcolsep}{8pt}
\resizebox{\linewidth}{!}{%
\begin{tabular}{lccc}
\toprule
\textbf{Dataset} & \textbf{MedVAE AE (PSNR)} & \textbf{SD-VAE AE (PSNR)} & \textbf{$\Delta$} \\
\midrule
MRNet & 27.85 & 23.92 & $+3.93$ \\
BraTS & 37.87 & 31.39 & $+6.48$ \\
CXR   & 36.93 & 32.02 & $+4.91$ \\
\bottomrule
\end{tabular}}
\end{table}

\begin{table}[h!]
\centering
\small
\caption{\textbf{Supplementary Table S12.} Fr\'{e}chet Inception Distance (FID) across all methods.
Lower is better. MRNet FID values are unreliable ($n=120 < 2{,}048$ minimum).}
\label{tab:fid}
\setlength{\tabcolsep}{5pt}
\resizebox{\linewidth}{!}{%
\begin{tabular}{@{}lcccccc@{}}
\toprule
\textbf{Dataset} & \textbf{MedVAE AE} & \textbf{MedVAE SR} & \textbf{SD-VAE SR} & \textbf{ESRGAN} & \textbf{SwinIR} & \textbf{Bicubic} \\
\midrule
MRNet$^\dagger$ & 46 & 76 & 73 & 158 & 221 & 161 \\
BraTS           & \textbf{9}  & 38 & 31 & 111 & 83  & 102 \\
CXR             & \textbf{15} & 59 & 48 & 101 & 74  & 111 \\
\bottomrule
\end{tabular}}

\smallskip
{\footnotesize $\dagger$ MRNet $n=120 < 2{,}048$; FID not reliable.}
\end{table}

\clearpage

\section{Supplementary Figures}

\begin{figure}[htbp]
  \centering
  \includegraphics[width=\linewidth]{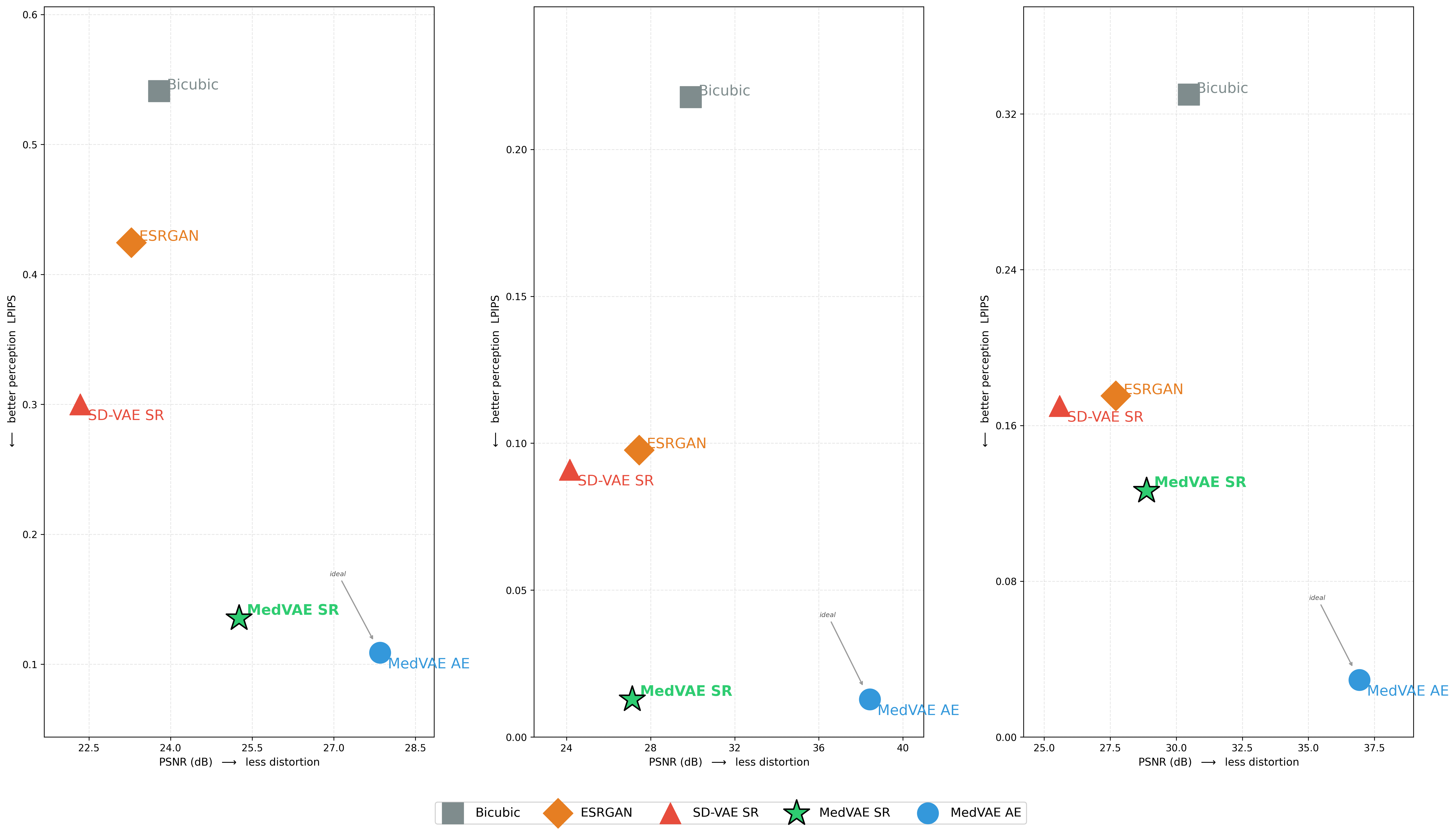}
  \caption{Perception-distortion tradeoff across three medical imaging datasets (MRNet, BraTS, MIMIC-CXR).
  Each point represents one SR method evaluated on one dataset.
  $x$-axis: PSNR (dB; higher = less distortion); $y$-axis: LPIPS (lower = more perceptually faithful).
  \medvaesr{} moves toward the upper-right Pareto frontier relative to \sdvaesr{}, achieving
  simultaneously higher PSNR and lower LPIPS on all three datasets.}
  \label{fig:supp_pd}
\end{figure}

\begin{figure}[htbp]
  \centering
  \includegraphics[width=0.62\linewidth]{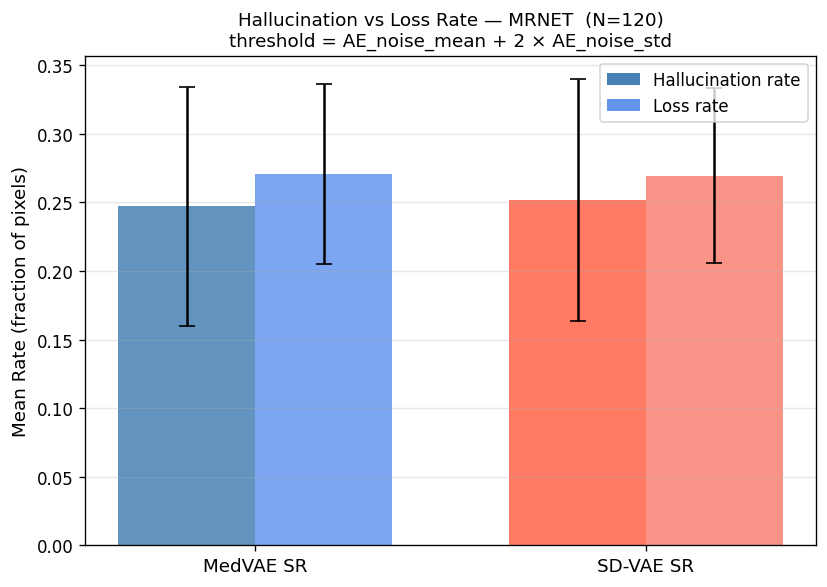}
  \caption{Pixel-level hallucination and content loss --- MRNet (knee MRI).
  Bar chart showing hallucination rate and loss rate (\% of pixels) for \medvaesr{} and \sdvaesr{}.}
  \label{fig:hallucination_mrnet}
\end{figure}

\begin{figure}[htbp]
  \centering
  \includegraphics[width=0.62\linewidth]{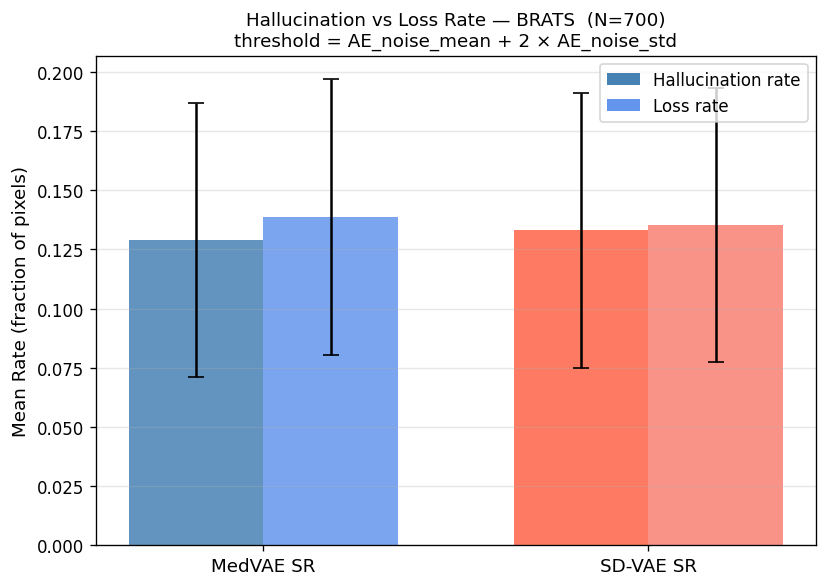}
  \caption{Pixel-level hallucination and content loss --- BraTS (brain MRI).}
  \label{fig:hallucination_brats}
\end{figure}

\clearpage

\begin{figure}[htbp]
  \centering
  \includegraphics[width=0.62\linewidth]{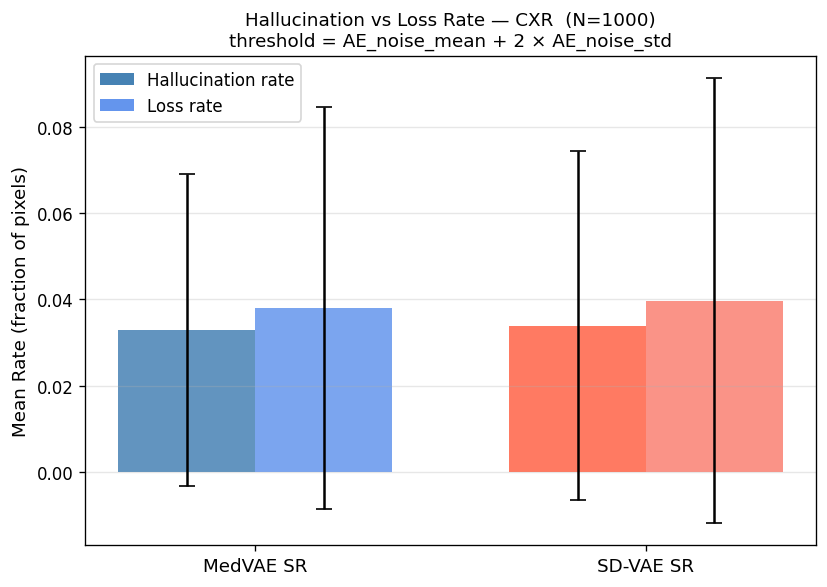}
  \caption{Pixel-level hallucination and content loss --- MIMIC-CXR (chest X-ray, $4\times$ SR).}
  \label{fig:hallucination_cxr}
\end{figure}

\clearpage

\section{Supplementary Methods}
\label{sec:suppmethods}

\paragraph{VAE architecture details.}
MedVAE (\texttt{medvae\_4\_3\_2d}) uses a KL-regularised convolutional autoencoder with output latent
shape $3 \times 64 \times 64$ and a compression factor of $4\times$.
The encoder uses residual blocks with group normalisation and SiLU activations.
SD-VAE (\texttt{stabilityai/sd-vae-ft-ema}) produces latents of shape $4 \times 32 \times 32$,
achieving $8\times$ spatial compression.
Both VAEs were used with frozen weights throughout all LDM experiments.

\paragraph{Dataset preprocessing.}
All images were resized to $256 \times 256$ pixels (bilinear interpolation) and normalised to $[0,1]$.
For $4\times$ super-resolution tasks (BraTS, MIMIC-CXR), low-resolution inputs were obtained
by bicubic downsampling to $64 \times 64$ followed by bicubic upsampling back to $256 \times 256$.
For the MRNet refinement task, mild Gaussian noise was added to the LR input during training.
No data augmentation was applied during evaluation.

\paragraph{Statistical methodology.}
Wilcoxon signed-rank tests were computed per-image on PSNR pairs (MedVAE SR vs.\ SD-VAE SR).
Cohen's $d$ effect sizes and 95\% bootstrap confidence intervals used $n_\mathrm{bootstrap} = 10{,}000$
resamples drawn with replacement from the per-image PSNR differences.
All statistical tests are two-sided; no correction for multiple comparisons was applied given the
a priori hypothesis tested (MedVAE SR $>$ SD-VAE SR).

\paragraph{Wavelet decomposition.}
Three-level Haar discrete wavelet transform (DWT) was applied to each grayscale image using \texttt{PyWavelets}.
Each image yields 10 subbands: one coarse approximation (LL$_3$) and three detail bands
(LH, HL, HH) at each of the three decomposition levels.
Per-subband PSNR was computed between SR and HR subband coefficients,
averaged across all validation images.

\paragraph{Hallucination quantification.}
For each SR image, the signed pixel difference $D = \text{SR} - \text{HR}$ was computed.
Using the corresponding AE image as a noise-floor baseline,
pixels satisfying $|D| > \mu_{|\mathrm{AE}-\mathrm{HR}|} + 2\sigma_{|\mathrm{AE}-\mathrm{HR}|}$
were classified as \emph{hallucinated} (if $D > 0$) or \emph{lost} (if $D < 0$).

\paragraph{Multi-resolution latent embedding.}
SR latents (shape $3 \times 64 \times 64$) were saved after the diffusion sampling step.
HR latents were obtained by passing HR images through the frozen MedVAE encoder.
Both SR and HR latents were average-pooled to progressively coarser resolutions using
\texttt{torch.nn.functional.avg\_pool2d}. Cosine similarity was computed channel-wise then averaged.

\section{Supplementary Notes}

\subsection{Supplementary Note S1: CXR Mid-Low Frequency Anomaly}
\label{sec:note1}

On MIMIC-CXR, \sdvaesr{} marginally outperforms \medvaesr{} at mid-low spatial frequency
subbands (LH$_2$/HL$_2$: $-0.61$ to $-0.81$~dB).
To determine whether this reflects a diffusion-stage or encoder-stage property,
we computed wavelet-subband PSNR on AE-only images.
The same pattern is present in the AE images: \sdvaeae{} scores higher at LH$_2$/HL$_2$
by $0.05$--$0.10$~dB.
This confirms that the anomaly is a property of the SD-VAE encoder's latent representation,
not an artefact of the downstream diffusion model.
A plausible explanation is that SD-VAE's larger channel capacity ($4 \times 32 \times 32$)
more efficiently encodes the broad, slowly-varying contrast gradients that dominate
chest radiograph mid-low frequency content.
At the finest scale (HH$_1$, fine detail), MedVAE retains its advantage
($+0.70$~dB on CXR), consistent with its domain-specific training on medical textures.

\subsection{Supplementary Note S2: MRNet Task Description}
\label{sec:note2}

The MRNet dataset~\citep{bien2018mrnet} comprises 1,370 knee MRI exams (axial slices,
$256\times256$ pixels) acquired at Stanford University Medical Center (2001--2012).
Unlike BraTS and MIMIC-CXR, where the LR input is a $4\times$ bicubic downsample of the HR,
the MRNet task is a \emph{refinement} task: the LR image is constructed by applying Gaussian
blur ($\sigma=1$), downsampling $2\times$, then upsampling back to $256\times256$.
Both LR and HR therefore share the same spatial dimensions ($256\times256$),
and the SR model must recover fine anatomical detail lost during the simulated degradation.

\begin{table}[h!]
\centering
\caption{\textbf{Supplementary Table S13.} Cosine similarity between SR latents and HR latents
across diffusion timestep budgets at the global mean ($1\times1$) scale.
The narrow range ($\Delta \leq 0.025$) confirms that latent fidelity is determined primarily by the VAE architecture.}
\label{tab:multit_embed}
\setlength{\tabcolsep}{7pt}
\resizebox{\linewidth}{!}{%
\begin{tabular}{lccccl}
\toprule
\textbf{Dataset} & $T=50$ & $T=100$ & $T=250$ & $T=1{,}000$ & $\Delta$ \\
\midrule
MRNet & 0.962 & 0.975 & 0.968 & 0.956 & 0.019 \\
BraTS & 0.971 & 0.984 & 0.978 & 0.963 & 0.021 \\
CXR   & 0.967 & 0.981 & 0.975 & 0.961 & 0.020 \\
\bottomrule
\end{tabular}}
\end{table}

\subsection{Supplementary Note S3: Disentangling Latent Capacity from Domain Specificity}

MedVAE employs a $3\times64\times64 = 12{,}288$-dimensional latent space, compared with SD-VAE's $4\times32\times32 = 4{,}096$ dimensions --- a $3\times$ difference in total capacity.
This is architecturally infeasible without retraining: the MedVAE UNet expects a 6-channel $64\times64$ input (3ch noisy latent + 3ch LR condition), whereas the SD-VAE UNet expects an 8-channel $32\times32$ input (4ch + 4ch).
The AE ceiling comparison in Appendix Table~\ref{tab:ae_ceiling} already provides the requested disentanglement: both AE evaluations bypass the UNet entirely (pure encode--decode), varying only the encoder/decoder while holding all downstream processing constant.
The SD-VAE AE ceilings are $23.92$~dB (MRNet), $31.39$~dB (BraTS), and $32.02$~dB (CXR); the MedVAE AE ceilings are $27.85$~dB, $37.87$~dB, and $36.93$~dB --- gaps of $+3.93$, $+6.48$, and $+4.91$~dB attributable entirely to encoder/decoder quality.
These AE gaps are consistent with and bound the SR gains ($+2.91$--$+3.29$~dB), providing strong evidence that domain specificity rather than latent capacity drives the improvement.

\subsection{Supplementary Note S4: Inference Efficiency at $T=16$ Steps}

We directly evaluated \medvaesr{} at $T=16$ DDPM steps across all three datasets.
$T=16$ achieves LPIPS of $0.218$ (MRNet), $0.086$ (BraTS), and $0.301$ (CXR), compared to $T=100$ LPIPS of $0.135$, $0.088$, and $0.127$ respectively.
The results reveal a clear perception-distortion tradeoff: $T=16$ produces outputs closer to the LR conditioning, while $T=100$ introduces greater perceptual detail at the cost of some pixel-level fidelity.
Progressive distillation~\citep{salimansProgressiveDistillationFast2022} offers a complementary route to inference acceleration and is a natural candidate for future integration with the MedVAE pipeline.

\section{Supplementary Figures (continued)}

\begin{figure}[htbp]
  \centering
  \includegraphics[width=\linewidth,height=0.42\textheight,keepaspectratio]{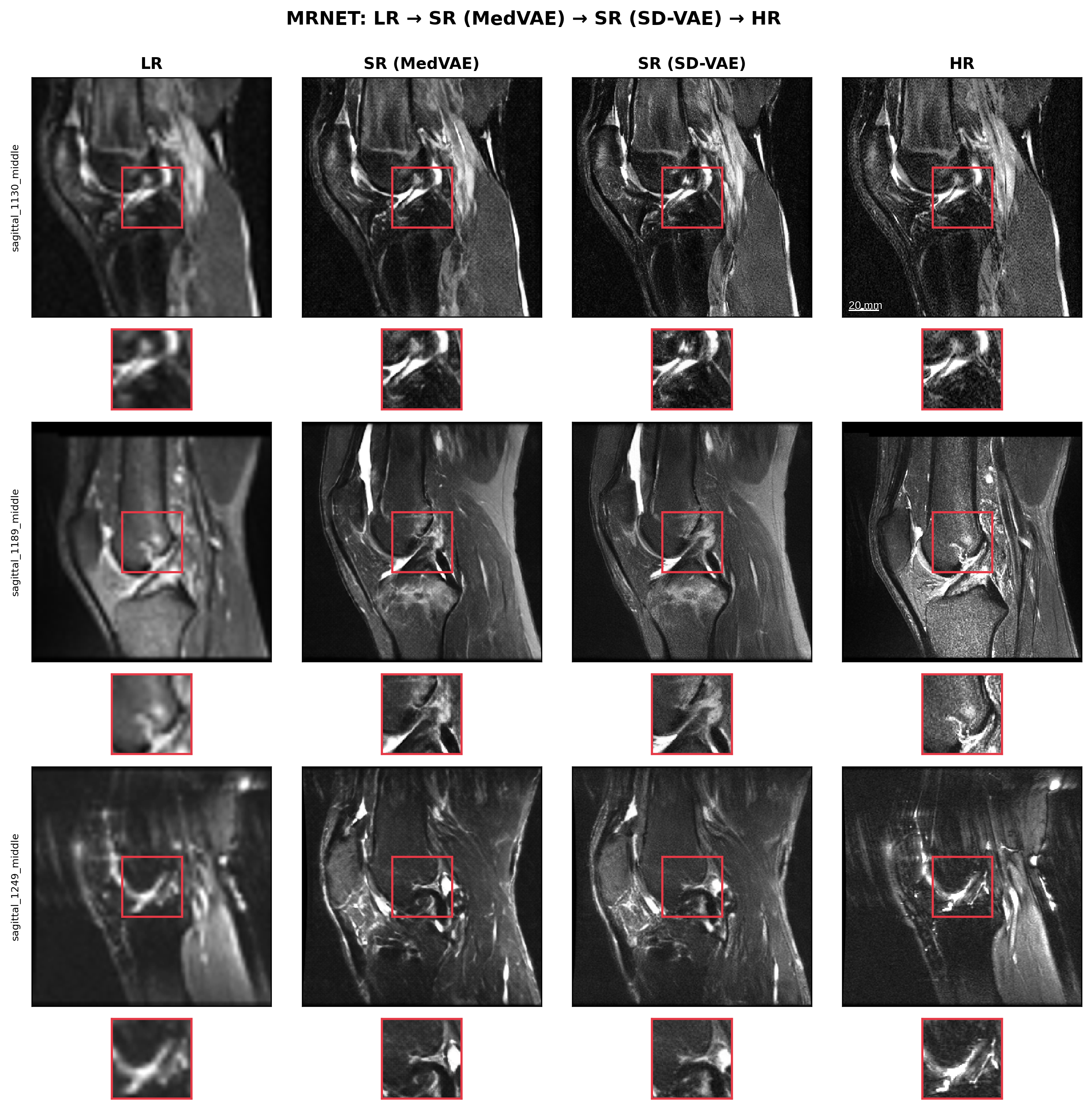}
  \vspace{0.2em}
  \includegraphics[width=\linewidth,height=0.18\textheight,keepaspectratio]{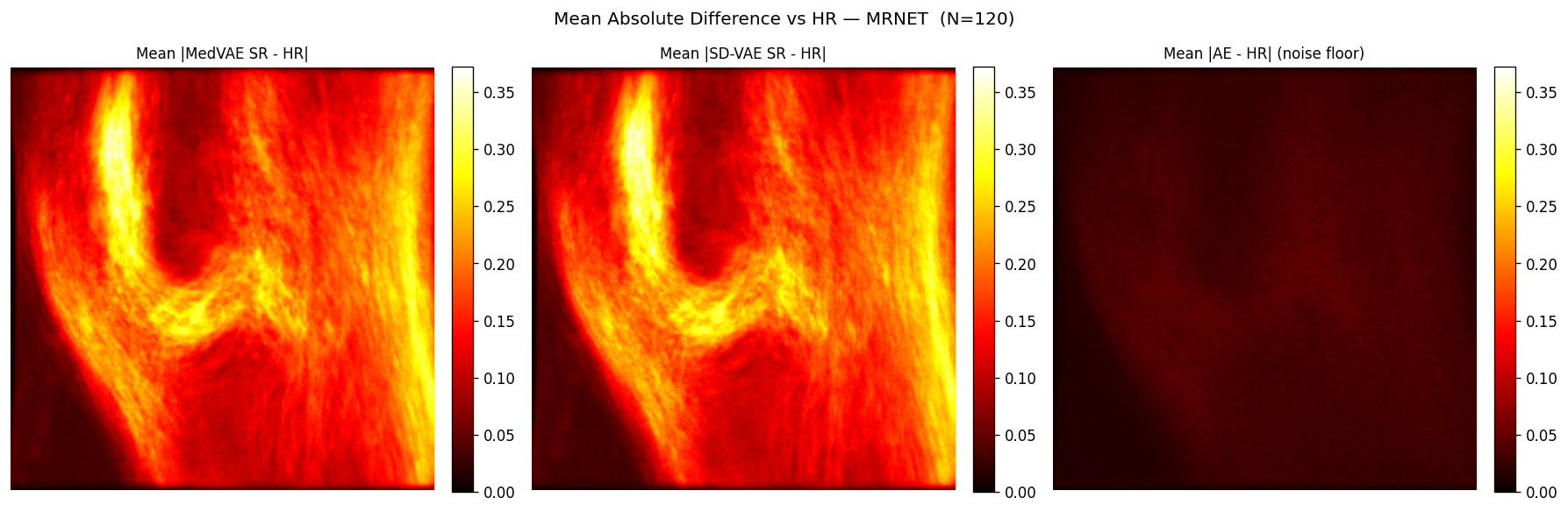}
  \caption{Visual comparison and pixel-level difference maps --- MRNet (knee MRI).
  Top: side-by-side comparison of AE reconstruction, \medvaesr{}, \sdvaesr{}, and HR ground truth.
  Bottom: mean absolute pixel difference $|\text{SR}-\text{HR}|$ (hot colormap; brighter $=$ larger error).}
  \label{fig:visual_mrnet}
\end{figure}

\clearpage

\begin{figure}[htbp]
  \centering
  \includegraphics[width=\linewidth,height=0.42\textheight,keepaspectratio]{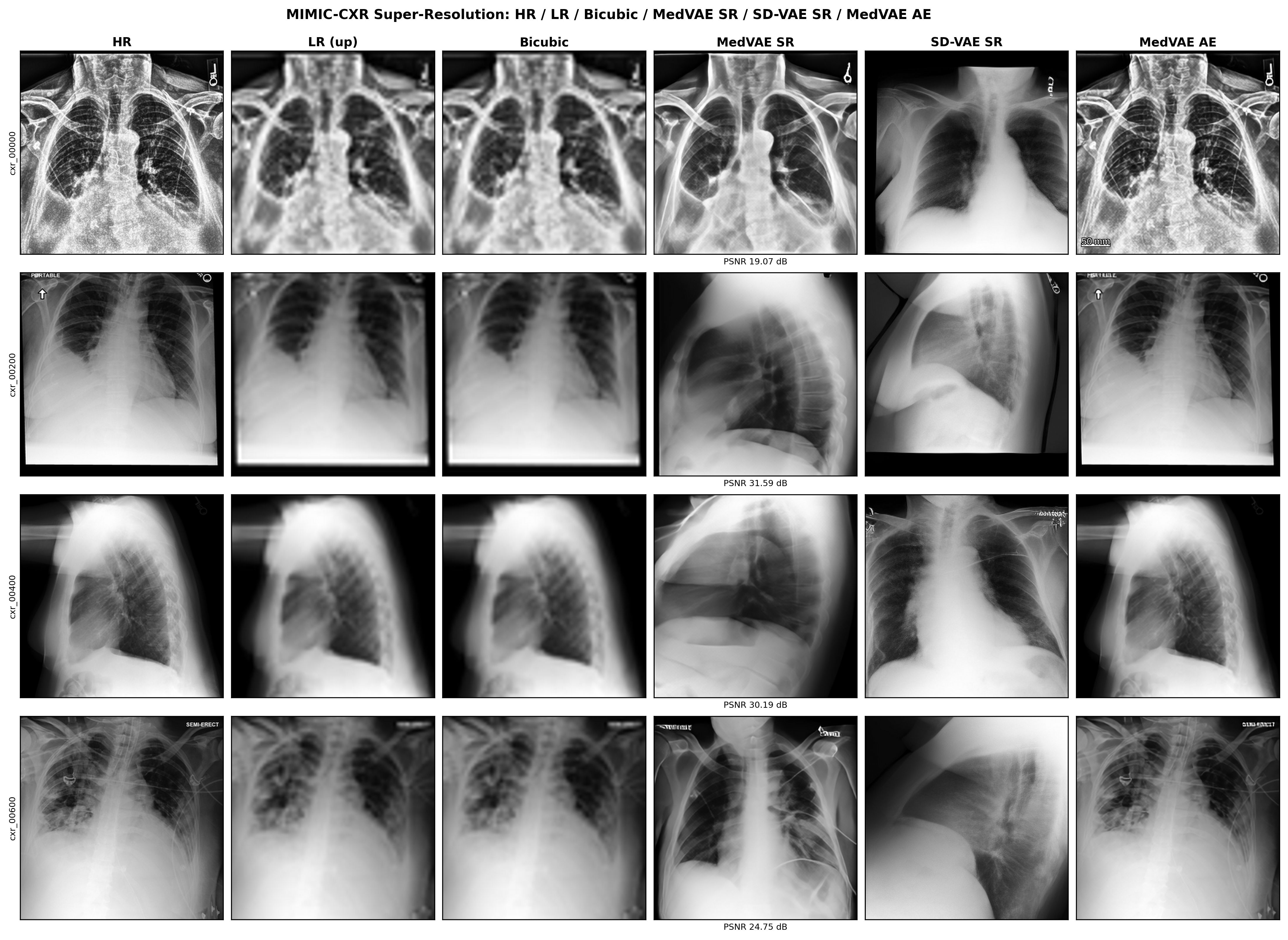}
  \vspace{0.2em}
  \includegraphics[width=\linewidth]{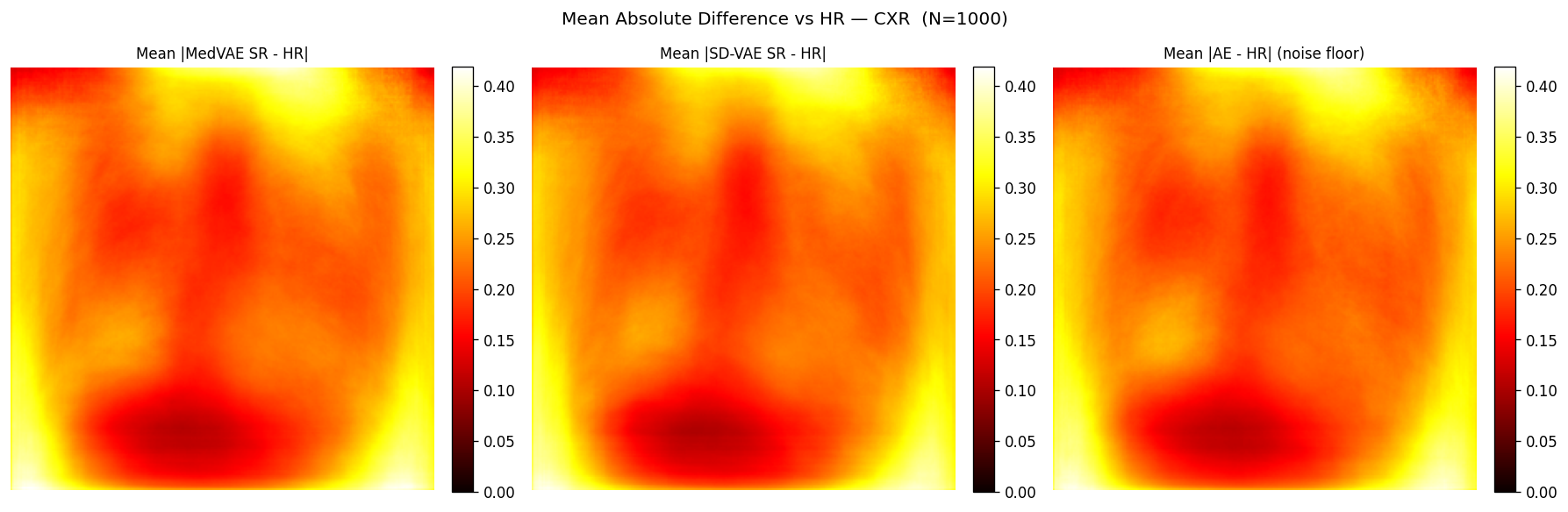}
  \caption{Visual comparison and pixel-level difference maps --- MIMIC-CXR (chest X-ray, $4\times$ SR).
  Top: side-by-side comparison of AE reconstruction, \medvaesr{}, \sdvaesr{}, and HR ground truth.
  Bottom: mean absolute pixel difference $|\text{SR}-\text{HR}|$ (hot colormap).}
  \label{fig:visual_cxr}
\end{figure}

\end{document}